\documentclass{article}
\usepackage{iclr2027_conference,times}

\usepackage{amsmath,amsfonts,bm}

\def\eqref#1{equation~\ref{#1}}

\def\1{\bm{1}}

\DeclareMathAlphabet{\mathsfit}{\encodingdefault}{\sfdefault}{m}{sl}
\SetMathAlphabet{\mathsfit}{bold}{\encodingdefault}{\sfdefault}{bx}{n}

\usepackage[colorlinks=true, linkcolor=black, citecolor=black,
            urlcolor=blue, breaklinks=true]{hyperref}
\usepackage{url}
\usepackage{array}      %
\usepackage{booktabs}
\usepackage{graphicx}
\usepackage{amsmath}
\usepackage{amssymb}
\usepackage{multirow}
\usepackage{subcaption}
\usepackage{xcolor}

\newcommand{\NAssays}{31}
\newcommand{\NOrganoidAssays}{18}
\newcommand{\NSliceAssays}{13}
\newcommand{\NOrganoidRoutings}{13}

\newcommand{\ChanMin}{841}
\newcommand{\ChanMax}{1{,}020}
\newcommand{\CorpusHours}{2.8}
\newcommand{\DurMinS}{127}
\newcommand{\DurMaxS}{1,940}

\newcommand{\CorpusMin}{166}
\newcommand{\OrganoidMin}{98}
\newcommand{\SliceMin}{68}

\newcommand{\FrameMs}{6}
\newcommand{\WindowMs}{600}
\newcommand{\ClipMs}{288}
\newcommand{\ClipFrames}{48}
\newcommand{\ClipVoxels}{1,290,240}
\newcommand{\NWindows}{2,133}
\newcommand{\VoxelRate}{$1.62\times 10^{-4}$}
\newcommand{\SpikesPerClip}{209}
\newcommand{\NTrain}{1069}
\newcommand{\NVal}{426}
\newcommand{\NTest}{638}
\newcommand{\WindowedMin}{21}
\newcommand{\TrainMin}{10.7}
\newcommand{\ValMin}{4.3}
\newcommand{\TestMin}{6.4}

\newcommand{\CeilingFactor}{5.2}
\newcommand{\OracleOurs}{0.2535}
\newcommand{\OracleFlat}{0.0486}
\newcommand{\OracleOursSite}{0.2296}
\newcommand{\OracleFlatSite}{0.2607}
\newcommand{\NTestClips}{279}

\newcommand{\NTaskSeed}{20260822}
\newcommand{\NTaskMC}{8}

\title{A discrete generative model of neuronal \\ spiking activity on microelectrode arrays}

\author{
  Md Sayed Tanveer$^{1,2}$ \;
  Mohammed A. Mostajo-Radji$^{3}$ \; Ge Wang$^{1,2,\ast}$ \\[0.4em]
  \normalfont\normalsize
  $^{1}$Department of Biomedical Engineering,
  Rensselaer Polytechnic Institute, Troy, NY, USA \\
  \normalfont\normalsize
  $^{2}$Center for Biotechnology and Interdisciplinary Studies,
  Rensselaer Polytechnic Institute \\
  \normalfont\normalsize
  $^{3}$Genomics Institute, University of California Santa Cruz,
  Santa Cruz, CA 95064, USA \\[0.3em]
  \normalfont\normalsize
  \texttt{azamm@rpi.edu} \quad \texttt{mmostajo@ucsc.edu} \quad
  \texttt{wangg6@rpi.edu} \\[0.2em]
  \normalfont\normalsize $^{\ast}$Corresponding author.
}

\iclrfinalcopy

\begin{document}
\maketitle
\lhead{}\renewcommand{\headrulewidth}{0pt}

\begin{abstract}
Generative models of neural activity could help characterize tissue dynamics,
compare experimental conditions, and simulate population activity for
applications ranging from disease and drug-response studies to closed-loop
experimentation. Existing approaches, however, typically assume a fixed set of
sorted neurons, whereas high-density microelectrode arrays produce extremely
sparse, array-wide binary spike volumes in which the observed subset of
electrodes varies across assays. We introduce a discrete generative model that
represents this activity using a shared vocabulary of spatiotemporal motifs. A
residual vector-quantized autoencoder learns the motif vocabulary, while a
factorized masked transformer predicts where activity occurs and which motif
appears at each active location. We evaluate the model on \NAssays\ assays
spanning human brain organoids and acute \emph{ex vivo} human hippocampal
tissue. The learned motifs are broadly reused: assay identity explains only
$9\%$ of the entropy in motif use, and motif overlap across tissue types is
comparable to overlap within them. When representation quality is evaluated
independently of the generative prior, our approach achieves
\CeilingFactor$\times$ the voxel-level reconstruction average precision of a
matched flat tokenizer. For masked completion and free generation, the full
model achieves $1.4$--$2.6\times$ the site-level average precision of the
matched generative baseline and outperforms it across all four families of
generation metrics.
These results establish a compact, reusable representation for array-wide
spiking activity without learned assay-specific parameters, providing a
scalable foundation for generative modeling across diverse neural preparations.
\end{abstract}

\section{Introduction}
\label{sec:intro}

Cultured neural tissue on high-density microelectrode arrays is being used in
closed-loop tasks \citep{kagan2022dishbrain, smirnova2023oi, patel2025neuroai,
robbins2026goaldirected}. A \emph{generative forward model} of spontaneous
activity could support two uses: simulate the preparation, and give the
unperturbed baseline against which stimulus-evoked change is measured.
Controller design needs stimulation-conditioned dynamics, absent from these
recordings (Section~\ref{sec:limitations}). Organoids develop rich spontaneous
population dynamics
\citep{trujillo2019oscillations, sharf2022organoids}, and HD-CMOS arrays record
them at single-electrode resolution across tens of thousands of sites
\citep{ballini2014hdmea}. Existing generative models of spiking activity work
on sorted units \citep{kapoor2024ldns, minnick2026spikeprophecy}; we sort too,
but write each unit to its peak electrode and model the array-wide volume with
no per-unit parameter. The activity is not uniform
noise; recurring population events have structured spatial and temporal
organization \citep{beggs2003avalanches, vandermolen2026preconfigured}. This
motivates asking whether such events can be represented by a shared motif
vocabulary.

Both preparations show structured population activity
\citep{sharf2022organoids, vandermolen2026preconfigured,
andrews2024hippocampal}. Per-assay learned tables can conflate reusable
structure with assay-specific memorization: removing assay identity strongly
degrades our statistical references (Section~\ref{sec:tasks}). We aim instead
for a shared model with no learned per-assay parameters.

Representing it is hard because of sparsity, not dimensionality. A clip is a $\ClipFrames\times120\times224$ binary volume with voxel occupancy
\VoxelRate, and only \ChanMin--\ChanMax\ of the array's $26{,}400$ sites are
routed in any assay, with a different subset each time. At this density the
structure is \emph{which electrodes participate together}, not image-like
intensity. We therefore evaluate both
site-level accuracy (which electrodes are active) and voxel-level accuracy
(which electrode is active in which frame).

\paragraph{Contributions.}
\begin{enumerate}
  \item \textbf{A motif alphabet for sparse spike volumes.} A three-level
        residual VQ-VAE learns spatiotemporal patches; its residual paths form
        one deduplicated categorical alphabet, with a dedicated blank token for
        empty patches. When both tokenizers are given the ground-truth codes for a
        held-out region, ours achieves $\CeilingFactor\times$ the voxel-level
        reconstruction AP of a flat-tokenizer baseline with the same grid and
        patch size (Section~\ref{sec:alphabet}).
  \item \textbf{Evidence that motifs are reused across observed assays.}
        Assay identity explains only $9\%$ of code entropy, and vocabulary
        overlap across preparation types is similar in magnitude to overlap
        within them (Section~\ref{sec:reuse}). This analysis is possible
        because the corpus spans cultured organoid tissue and acute \emph{ex
        vivo} human hippocampus.
  \item \textbf{A factorized prior with controlled comparisons.} The prior
        predicts first where activity occurs and then which motif occupies each
        active location. It is conditioned on a fixed per-assay code but
        adds no learned assay-specific parameter table, so assay identity remains
        available while the number of learned parameters remains independent
        of the number of assays. We compare it with a
        same-grid flat-tokenizer baseline, directly supervised convolutional
        models, and assay-specific statistical references. Oracle decoding
        identifies temporal prediction as the principal bottleneck ($80$--$98\%$
        of site-level versus $4$--$6\%$ of voxel-level representational
        capacity) (Section~\ref{sec:experiments}).
\end{enumerate}

\section{Related work}
\label{sec:related}

The dichotomized Gaussian \citep{macke2009dg} and the coupled point-process GLM
\citep{pillow2008glm, truccolo2005pointprocess} are the standard forward models
for binary population activity, and both are fitted \emph{per assay}: their
parameters are a rate vector and a coupling or covariance matrix over that
assay's electrodes. Those spatial parameters scale with the number of
assays, so we include both as assay-specific statistical references and
not as parameter-matched peers (Section~\ref{sec:tasks}). Sparsity limits
what can be fitted, so each uses the stationary parameterization its
literature prescribes (Appendix~\ref{app:baselines}).

Our own construction borrows from discrete generative modeling of images and
video. Vector-quantized autoencoders \citep{vandenoord2017vqvae,
razavi2019vqvae2, esser2021vqgan} turn continuous signals into token grids that
an autoregressive or masked prior can model, and MaskGIT's parallel iterative
unmasking \citep{chang2022maskgit} extended this to video in MAGVIT
\citep{yu2023magvit, yu2024magvit2}. Residual quantization is prior art from
image and audio coding \citep{lee2022rqvae, zeghidour2022soundstream}; we use
it without claiming it as a contribution. Two adaptations make the residual
ladder usable by a masked prior: a sparse encoder for a canvas that is $89\%$
empty, and flattening the ladder into one deduplicated categorical alphabet so
the prior predicts one symbol per site. Our peer
method, \textbf{MaskGIT-flat}, is the ordinary version of that construction at
a matched token budget, and the oracle-code comparison in
Section~\ref{sec:alphabet} takes the prior out of it.

The tissue and the instrument bring their own literature. Human brain organoids
develop oscillatory and avalanche-structured spontaneous activity
\citep{trujillo2019oscillations, beggs2003avalanches}, and HD-CMOS arrays
resolve it at single-electrode scale \citep{ballini2014hdmea}. The organoid
recordings were produced by \citet{sharf2022organoids}. Later work combined
them with further recordings and reported preconfigured, repeating population
sequences \citep{vandermolen2026preconfigured}. That literature
characterizes such sequences; we learn an alphabet of them and build a
generative model on the two deposits \citep{dandi000732,
dandi001132}, with spike detection and sorting following standard practice
\citep{buccino2020spikeinterface}.

Closest in aim are the generative models for neural recordings. LFADS
\citep{pandarinath2018lfads} and transformer successors \citep{ye2021ndt} infer
latent dynamics underlying population activity; Spike-GAN
\citep{molano2018spikegan} generates spike trains adversarially, and LDNS
\citep{kapoor2024ldns} does so with a latent diffusion model conditioned on
behavior. All operate on tens to a few hundred \emph{sorted units}. Models
built to span sessions exist: POYO \citep{azabou2023poyo} tokenizes individual
units and NDT2 \citep{ye2023ndt2} pretrains across sessions and subjects, but
both learn a per-unit or per-session embedding, so their parameters still grow
with the recordings covered. We model the array-level
binary volume directly on the $120\times224$ canvas, with no unit-level
parameterization and no learned per-assay table, which is what the reuse
measurement in Section~\ref{sec:reuse} justifies. A second recent line forecasts population
spiking autoregressively at Neuropixels scale: a state-space forecaster trained
on next-step spike counts also supports a linear behavioral readout
\citep{minnick2026implicit}, and a companion benchmark separates the usual
aggregate correlation into temporal fidelity, spatial pattern accuracy and
magnitude-invariant alignment \citep{minnick2026spikeprophecy}. Both work on
the next bin of sorted-unit counts, while we complete masked regions of a
binary array volume, so neither their readout nor their metric transfers
directly. A third line generates spikes for a downstream target:
\citet{wu2026generative} map sorted upstream neurons to firing probabilities
for a small number of downstream neurons and sample spikes from those
probabilities. Their generator is optimized by behavioral reward rather than
direct downstream-spike supervision, although the behavioral decoder supplying
that reward is itself trained on recorded downstream activity. Their objective
is task-directed communication between two regions. All of this is adjacent
work, addressing sorted-unit latent dynamics, population generation, or
next-bin and transregional prediction, not the array-wide binary volume.

\section{Data and task}
\label{sec:data}

We use \NAssays\ recordings from two open-access DANDI dandisets:
\NOrganoidAssays\ of human brain organoid slices \citep{dandi000732} and
\NSliceAssays\ of human \emph{ex vivo} hippocampal slices resected during
neurosurgery \citep{dandi001132}. Together, they contain \CorpusMin\ minutes of
recording: \OrganoidMin\ minutes from organoid slices and \SliceMin\ minutes
from hippocampal slices, in files that run from two minutes to half an hour
(Appendix~\ref{app:data}). Both datasets were recorded using the same MaxWell
Biosystems high-density CMOS array. The array contains $26{,}400$ sites at a
pitch of $17.5\,\mu\mathrm{m}$, of which $1{,}024$ can be read simultaneously
at $20$\,kHz. Per assay,
\ChanMin--\ChanMax\ channels are routed, i.e.\ under $4\%$ of the array, and a
different subset each time. The model's canvas is therefore the array footprint
itself, $120\times220$ sites padded to $120\times224$: routing varies between
assays, physical coordinates do not, so one shared model spans every routing
configuration. Both preparation types are included because an alphabet learned
from cultured organoid tissue cannot be assumed to represent acute resected
hippocampal tissue.

We treat each archive file as one \emph{assay} with an assay ID, the unit of
conditioning in the model. An assay ID is a dataset identifier and not a
biological one: several assays can come from one organoid or one tissue
preparation, recorded in different sessions or under different experimental
conditions. The \NOrganoidAssays\ organoid assays come from a source dataset of
array recordings of six sectioned human brain organoids, L1--L6, grown from one
donor-derived induced pluripotent stem cell line. The source dataset includes
repeated recordings across developmental ages and drug conditions
\citep{sharf2022organoids}. The
conversion to NWB preserves neither the organoid nor the condition behind any
assay, and three pairs of organoid assay files hold the same acquisition trace
(Appendix~\ref{app:data}). The \NSliceAssays\ slice assays record two slice
preparations, from a woman of $52$ and a man of $35$, whose tissue was removed
during temporal lobectomy for drug-refractory epilepsy
\citep{andrews2024hippocampal}. All splits are temporal within an assay and none
is cross-preparation (Section~\ref{sec:limitations}).

Spikes are detected and sorted per assay with SpyKING Circus~2
\citep{buccino2020spikeinterface} and each curated unit is written to its peak
electrode as a point event. Activity is segmented into \NWindows\ burst windows
of \WindowMs\,ms and max-pooled into \FrameMs\,ms frames, so a frame records
\emph{whether} an electrode fired and not how many times. We distinguish three
durations. The recordings run for \CorpusMin\ minutes, the burst windows
cut out of them hold \WindowedMin\ minutes of activity, and the temporal split
of those windows inside each recording gives \NTrain/\NVal/\NTest\ windows,
which is \TrainMin/\ValMin/\TestMin\ minutes of training, validation and test
material (Appendix~\ref{app:preproc}). A random contiguous span of each window
is trimmed to \ClipFrames\ frames.

A clip is a binary volume $X \in \{0,1\}^{T \times H \times W}$ with
$T = \ClipFrames$, $H = 120$, $W = 224$: \ClipVoxels\ voxels at occupancy
\VoxelRate, about \SpikesPerClip\ spikes in \ClipMs\,ms. It is cut into
$(6,15,14)$ patches, producing a token grid of $8\times8\times16 = 1024$ sites. We
write $Z$ for the token field. It factors into an activity field $A$, which
marks which sites are non-empty, and a motif field $M_Z$, which names what
occupies each active one. Conditioning is a per-assay code $g_r$ and a
per-clip descriptor $\ell(X)$ (Section~\ref{sec:conditioning-def}).

A task is a pair $(q, M)$: an index $q \in \{0,1,2,3\}$ and the binary mask
$M$ over the grid marking what is hidden. Each task is sampled with probability
$\tfrac14$ and evaluated separately. \emph{Free generation} ($q = 0$) hides
everything, so it has no visible-volume context and is not a completion task.
\emph{Causal} completion is forward prediction from a kept prefix of
$25$--$75\%$ of frames. \emph{Noncausal} completion interpolates from both
sides, hiding a contiguous $30\%$ of frames. \emph{Spatial} completion imputes
an unobserved region of the array, a box covering $25$--$60\%$ of
$H \times W$ at an aspect ratio between $0.5$ and $2$, hidden across every
frame. Holes are snapped to the patch lattice before scoring. The four settings
hide different fractions of a clip, so base rates differ and comparisons hold
\emph{within} a setting only (Appendix~\ref{app:metrics}).

One term is easily confused with free generation. \emph{Reconstruction} refers
only to the tokenizer path (encode a true clip, quantize and decode), and it
measures what the alphabet can represent, independently of the prior.

\section{Method}
\label{sec:method}

\begin{figure}[t]
  \centering
  \includegraphics[width=\textwidth]{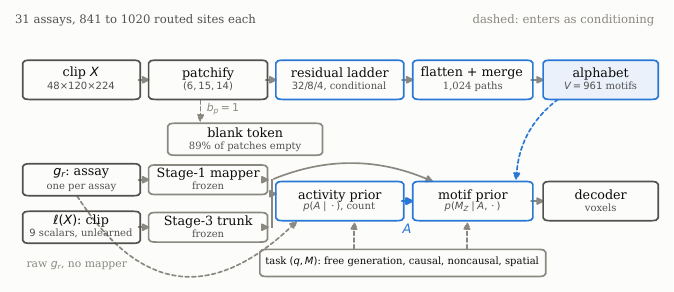}
  \caption{The pipeline. Empty patches bypass the quantizer; the three-level
  residual ladder is flattened to $V=961$ motifs. Assay and clip context
  condition the activity and motif priors by the routes defined in
  Section~\ref{sec:conditioning-def}, and the motif prior additionally reads the
  activity field emitted by the first prior.}
  \label{fig:pipeline}
\end{figure}

A tokenizer learns a shared alphabet of spatiotemporal motifs, after which a factorized prior models their arrangement (Figure~\ref{fig:pipeline}). The main text defines each objective
and constraint family; coefficients and curricula are in
Appendix~\ref{app:objectives}.

\subsection{A tokenizer with a route around the quantizer}
\label{sec:tokenizer}

A convolutional stem embeds patches, a $2$-layer transformer of width $64$ with
$4$ heads attends over the $8\times8\times16$ grid, and a
transposed-convolution renderer maps each token back to its $1260$ voxels.
Convolution supplies local feature extraction; global attention can relate
co-participating electrodes that are not spatial neighbors. We did not train a
matched all-convolutional core, so this choice is not isolated experimentally
(Section~\ref{sec:limitations}).

At this occupancy, $88.9\%$ of patches contain no spike. The encoder sets
$b_p=1$ for an empty patch and routes it to a dedicated blank embedding outside
the content quantizers. In our trained tokenizer $91.7\%$ of tokens take this
route; under an otherwise matched dense design, blank patches consume level-1
content codes (Section~\ref{sec:stages}). Patch size is likewise chosen for the
tokenizer--prior pair: smaller patches reconstruct better but produce a larger,
more blank-dominated token grid.

\subsection{Residual hierarchy and flattened alphabet}
\label{sec:flatten}

Content patches use a three-level residual ladder
\citep{lee2022rqvae,zeghidour2022soundstream} over a VQ-VAE
\citep{vandenoord2017vqvae}. For encoder output $h$,
\begin{equation}
  z_1=Q_1(h),\quad
  z_2=Q_2(h-z_1\mid z_1),\quad
  z_3=Q_3(h-z_1-z_2\mid z_1,z_2),\quad
  \tilde z=z_1+z_2+z_3,
  \label{eq:ladder}
\end{equation}
with $32$, $8$, and $4$ entries per level. Each level searches only children of
the previously selected path, giving $32\times8\times4=1024$ leaves. Level~1
captures the coarse motif and deeper levels refine its residual; cumulative
decoding shows that both refinements improve the rendered spike volume
(Section~\ref{sec:stages}). Codebooks are EMA buffers rather than parameters
updated by reconstruction gradients. Section~\ref{sec:alphabet} compares this
ladder with a flat codebook at the same token budget. When both tokenizers are
given their true codes, ours achieves \CeilingFactor$\times$ the voxel-level
reconstruction AP of the flat baseline.

The masked prior requires one categorical variable per token. Each residual path
is therefore materialized as the sum of its three code vectors; near-identical
sums are merged, leaving $V=961$ motifs (Appendix~\ref{app:arch}). The tokenizer
objective is
\begin{equation}
\mathcal L_{\rm tok}=\mathcal L_{\rm rec}+\mathcal L_{\rm VQ}
 +\lambda_c\mathcal L_{\rm ctx}+\lambda_f\mathcal L_{\rm field}
 +\lambda_s\mathcal L_{\rm spatial}+\lambda_g\mathcal L_{\rm gap}
 +\mathcal R_{\rm enc/blank}.
\label{eq:tokloss}
\end{equation}
Here $\mathcal L_{\rm rec}$ is tolerance-aware sparse-spike reconstruction and
$\mathcal L_{\rm VQ}$ is commitment/usage regularization. The remaining terms
match clip/patch descriptors, assay support and same-site short-gap
statistics; $\mathcal R_{\rm enc/blank}$ collects code-norm, blank-separation and
encoder anti-collapse constraints. Appendix~\ref{app:objectives} defines every
constituent term.

\subsection{Global and local conditioning}
\label{sec:conditioning-def}

Conditioning enters at two scales. Each assay ID has a fixed, seeded,
$L_2$-normalized $64$-D $\pm1$ code $g_r$, a dataset-level identifier and not a
measured biological descriptor or learned embedding table. Stage~1 learns a
$32$-D representation $c_g=f_g(g_r)$ by predicting the
assay's running union of active sites, minimizing
$\mathcal L_{\rm global}=\mathcal L_{\rm supp}^{\rm token}
+\mathcal L_{\rm supp}^{\rm voxel}+\lambda_{\rm sep}\mathcal L_{\rm sep}
+\lambda_{\rm adj}\mathcal L_{\rm adj}$. The support terms cover the measured
support while penalizing mass outside it; separation distinguishes assays
whose support maps differ, and adjacency matches assay-level short-gap
rates. The mapper is then frozen. The motif prior receives $c_g$, whereas the
activity prior uses a learned projection of the raw $g_r$.

The local descriptor $\ell(X)$ contains nine unlearned scalars: log mean firing
density; mass-weighted second moments in $x,y,t$; their three cross-terms; active
site ratio; and a temporal trend score. On completion tasks it is computed from
the whole clip, hidden region included, and handed over as a prompt: those
tasks measure context-controlled completion, not prediction from visible
activity alone. Stage~3 learns
$c_\ell=f_\ell(\ell(X))$ with
$\mathcal L_{\rm local}=\operatorname{NLL}_{\rm texton}
+\operatorname{NLL}_{\rm flat}+\mathcal L_{\rm summary}$, matching a
$128$-texton distribution, the $V=961$ flat-code histogram and per-code temporal
summaries. The texton basis is built on level~1 because the ablation shows that
these low-dimensional descriptors predict coarse motif identity better than the
residual refinements (Section~\ref{sec:stages}). After Stage~3, $c_\ell$ is
frozen and conditions both priors. Neither context changes the VQ lookup itself:
during tokenizer training it acts through decoded-output constraints, and during
generation it conditions selection over the frozen alphabet.

\subsection{A prior factorized into where and what}
\label{sec:prior}

A single categorical head would have to model the overwhelming blank mass and
the diversity of active motifs simultaneously. We instead write
\begin{equation}
  p(Z\mid g_r,\ell,q,M)=
  \underbrace{p(A\mid P_A g_r,c_\ell,q,M)}_{\text{where/how much}}
  \;\underbrace{p(M_Z\mid A,c_g,c_\ell,q,M)}_{\text{which motif}},
  \label{eq:factorization}
\end{equation}
where $P_A$ is the activity prior's learned projection of the raw assay code,
$c_g=f_g(g_r)$ and $c_\ell=f_\ell(\ell(X))$. Both factors are MaskGIT
transformers \citep{chang2022maskgit,yu2023magvit} with width $128$, four
layers, and four heads.

The \textbf{activity prior} emits a Bernoulli logit for each of the $1024$ token
cells plus a categorical head for total active count and reads the ROI occupancy
of $16$ coarse regions. Its objective is
\begin{equation}
\mathcal L_A=\mathcal L_{\rm cell}+\lambda_K\mathcal L_{\rm count}
+\lambda_t\mathcal L_{t\text{-coact}}+\lambda_x\mathcal L_{x\text{-coact}}
+\lambda_{\rm sup}\mathcal L_{\rm support}.
\label{eq:activityloss}
\end{equation}
This separates the two sparse-data failure modes. Aggressive class weighting
over-produces activity, while insufficient positive pressure collapses toward
blank. Our per-cell BCE is unweighted, and the count, co-activation
and support terms supply calibration and structure.

The \textbf{motif prior} iteratively places one of the $V=961$ motifs at each
active cell. Its objective is
\begin{equation}
\begin{aligned}
\mathcal L_M={}&\mathcal L_{\rm CE}+\lambda_n\mathcal L_{\rm nbr}
+\lambda_d\mathcal L_{\rm dist}+\lambda_c\mathcal L_{\rm ctx}
+\lambda_f\mathcal L_{\rm field}\\[-1mm]
&+\lambda_a\mathcal L_{\rm adj}+\lambda_s\mathcal L_{\rm spatial}.
\end{aligned}
\label{eq:motifloss}
\end{equation}
Exact CE selects the target motif; neighborhood CE and expected code distance
give graded credit to nearby entries in the frozen alphabet. The remaining
terms soft-decode logits through the frozen tokenizer so clip/patch context,
short-gap rates and assay support constrain the generated voxel field.
Coefficients are per objective, not shared: $\lambda_c$, $\lambda_f$ and
$\lambda_s$ name analogous constraints in Eq.~\ref{eq:tokloss} and
Eq.~\ref{eq:motifloss} but take different values, all of which are in
Appendix~\ref{app:objectives}.

\subsection{Staged training and inference-time adaptation}
\label{sec:training}

The final adaptation corrects the mismatch between teacher-forced motif
training and inference. Stage~4A learns motifs given the true activity map,
whereas at generation the motif prior receives the activity field emitted by
Stage~4B, whose cell-level recall is $0.72$. A missed active cell is an input
configuration absent under teacher forcing, not a noisier version of a seen
one. Stage~4C freezes the activity prior and fine-tunes the motif prior on
emitted maps. Table~\ref{tab:stages} gives the order and what each stage is
allowed to learn. The emitted field remains soft inside the hole so uncertainty
can be marginalized; replacing it with a hard map is worse than no adaptation
(Section~\ref{sec:stages}).

\begin{table}[t]
  \centering
  \small
  \begin{tabular}{@{}l >{\raggedright\arraybackslash}p{0.24\textwidth} >{\raggedright\arraybackslash}p{0.30\textwidth} >{\raggedright\arraybackslash}p{0.26\textwidth}@{}}
\toprule
stage & what it trains & supervision target & frozen output used later \\
\midrule
1 & global embedder + spatial-map head & the assay's running union of active sites & frozen code-to-support mapper \\
2A & encoder, decoder, EMA codebooks & the clip itself, reconstructed through the quantizer & frozen motif alphabet, $V=961$ \\
3 & lct trunk + heads (tokenizer frozen) & the clip's texton usage histogram & frozen nine-scalar-to-code mapper \\
4A & motif prior (tokenizer and mappers frozen) & true motifs, given the \emph{true} activity field & motif generator \\
4B & activity prior & the true activity field and its total count & activity generator \\
4C & motif prior (activity prior frozen) & true motifs, given the activity field 4B \emph{emits} & the prior used at inference \\
\bottomrule
\end{tabular}

\vspace{0.3em}

{\footnotesize Checkpoint selection: 1 on spatial-map loss; 2A on val exact AUPRC; 3 on val multinomial NLL; 4A on val MRR; 4B on val NLL; 4C on val MRR.}

  \caption{Training chronology. Each stage is frozen before the next begins.
  The supervision column identifies what each stage learns; full schedules and
  checkpoint criteria are in Appendix~\ref{app:training}.}
  \label{tab:stages}
\end{table}

\section{Experiments}
\label{sec:experiments}

Task-axis comparisons use paired Wilcoxon signed-rank tests with
Benjamini--Hochberg correction across the family \citep{benjamini1995fdr}; we
report $q$ values and identify other analyses separately. Each conditioning
result is compared with that model's random-context control; equal scores
indicate unused conditioning, except on adherence, which scores obedience to
the descriptor supplied. Average precision is stepwise, since trapezoidal
interpolation is invalid in PR space \citep{davis2006prcurves}. All results use
a fixed assay-balanced sample of the test split: \NTestClips\ clips, nine per
assay across all \NAssays\ assays (Appendix~\ref{app:metrics}).

\paragraph{Baselines.}
We compare our method with three learned models and two statistical references.
\textbf{MaskGIT-flat} is the matched peer, the ordinary construction this
literature describes: a fully convolutional VQ tokenizer with one flat codebook
of $1024$ entries, compared with our deduplicated alphabet of $961$, on the same grid and patch
size at a matched token budget. Core, codebook and prior all differ, so it
evaluates the combined design rather than any single component
(Appendix~\ref{app:baselines}).
\textbf{3D U-Net} is a direct-supervision reference, marked $\dagger$: an \emph{inpainter} trained on this exact hole distribution, so it cannot reconstruct. \textbf{3D CVAE} adds a conditional latent to that backbone, and that latent
collapses (Appendix~\ref{app:baselines}). The \textbf{dichotomized Gaussian} \citep{macke2009dg} and
\textbf{coupled GLM} \citep{pillow2008glm, truccolo2005pointprocess} are fitted
per assay as memorization references, marked \emph{ref}
(Appendix~\ref{app:baselines}).

\subsection{The alphabet}
\label{sec:alphabet}

\begin{figure}[t]
  \centering
  \includegraphics[width=\textwidth]{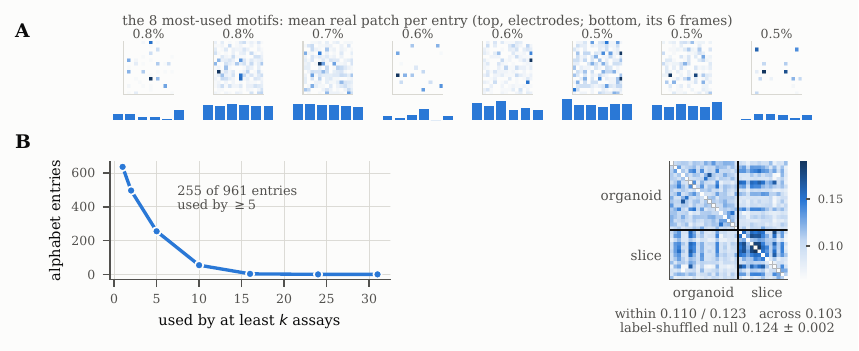}
  \caption{\textbf{What a motif represents, and whether it is reused.}
  \textbf{(A)} The most-used entries, each the mean \emph{real} voxel patch
  assigned to it, not a decoder rendering; time collapsed above, six frames
  below. \textbf{(B)} Entries used by at least $k$ assays, and pairwise
  vocabulary overlap by preparation type, both rarefied.}
  \label{fig:motifs}
\end{figure}

Figure~\ref{fig:motifs} shows what the entries represent and how broadly they
are reused. Reconstruction measures representational capacity and is compared
only with the other tokenizer.
Ours reaches step-wise AP $0.2564$ against MaskGIT-flat's $0.0269$. It does so
while using $619$ of its $961$ entries at perplexity $436.9$, against $844$
entries at $312.2$: fewer entries carrying more effective diversity.

Removing the prior isolates the representation. When each tokenizer is given
the \emph{true} codes for a hole, AP is \OracleOurs\ against \OracleFlat, a factor
of \CeilingFactor. The advantage is not uniform across the two axes: at
\emph{site} level ours is slightly \emph{lower}, \OracleOursSite\ against
\OracleFlatSite, so the hierarchical alphabet improves resolution in
\emph{time} rather than in space. Section~\ref{sec:tasks} shows that this is where the two arms
separate.

\subsection{Motifs are reused across assays}
\label{sec:reuse}

To test whether the alphabet is truly shared rather than partitioned by assay,
we report three statistics in Table~\ref{tab:reuse}.
$846$ of $961$ entries are in use across the test split, and $255$ of them are used by at least five of the \NAssays\ assays at a matched token budget. Knowing which assay a clip came from removes only $9\%$ of the code entropy.

The third statistic is why the corpus spans two preparation types. At matched
sampling effort, mean pairwise overlap within organoid assays is $0.1095$,
within slice assays $0.1232$, and \emph{across} the two types $0.1033$, so the same motifs appear in cultured organoid tissue and in acute human
hippocampal slices. Overlap sits slightly below the label-shuffle null, so a
small assay-specific component exists. These results do not establish transfer
to unseen assays or preparations
(Appendix~\ref{app:reuse}).

\subsection{Task completion}
\label{sec:tasks}

Full results for each setting and the corresponding per-clip comparisons are in
Appendix~\ref{app:battery} and Figure~\ref{fig:tasks}. At site level we achieve
$1.4$--$2.6\times$ the peer's AP in all four settings, winning $64$--$88\%$ of
clips ($q \leq 6.05\times10^{-10}$). At voxel level the comparison splits: we win
\emph{spatial} ($q = 8.21\times10^{-13}$) and \emph{free generation}
($q = 3.79\times10^{-4}$), while \emph{causal} and \emph{noncausal} are not
significant.

The oracle-code decomposition localizes this. Our prior recovers $80$--$98\%$
of its alphabet's site-level reference but only $4$--$6\%$ of its voxel-level
reference, MaskGIT-flat $44$--$57\%$ and $16$--$30\%$. Our binding constraint is
the prior's timing, theirs the flat alphabet.

Two reference classes outperform us. A static per-assay site map beats all
four learned models at both levels in every setting
($q \leq 3.1\times10^{-10}$), despite carrying no clip-specific information.
Its site-level AP falls from $0.6796$ to $0.0284$ when the clip's assay is
withheld, as do the two lookup references. Both directly supervised
convolutional models also beat us on
voxel AP in all four settings, the U-Net by the largest margin, and neither
holds a per-assay table. Direct supervision remains stronger for voxel-level
imputation; our contribution is a reusable discrete representation, which an
inpainter with an uncompressed skip path does not produce. Every arm above ours
across the $32$ paired comparisons is in one of these two classes, and none is
the matched peer (Appendix~\ref{app:baselines}).

\subsection{Generation}
\label{sec:generation}

\begin{figure}[t]
  \centering
  \includegraphics[width=\textwidth]{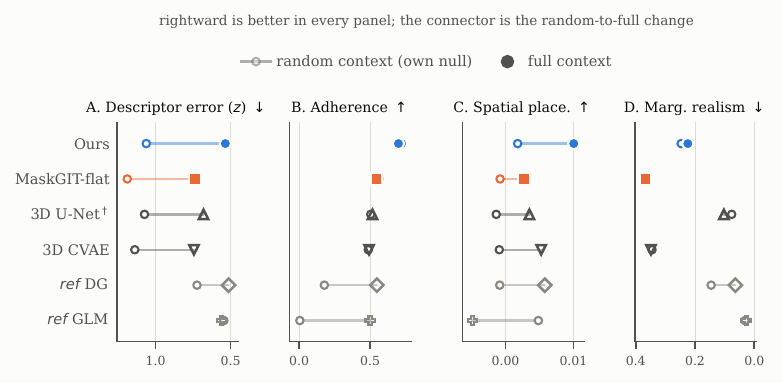}
  \caption{\textbf{Effect of conditioning for each model.} Hollow marker, that
  model's own random-context null; filled marker, full context; the connector is
  the random-to-full change. Coinciding markers mean unused conditioning in A, C
  and D; B scores obedience to the descriptor supplied, so a flat B is expected
  (Table~\ref{tab:fullladders}). Panels are oriented so rightward is better and have independent
  $x$ axes. All five context rungs appear in Table~\ref{tab:fullladders}.}
  \label{fig:generation}
\end{figure}

\begin{table}[t]
  \centering
  \small
  \resizebox{\textwidth}{!}{\begin{tabular}{l r r r r r r}
\toprule
family & Ours & MaskGIT-flat & 3D U-Net$^\dagger$ & 3D CVAE & \emph{ref} DG & \emph{ref} GLM \\
\midrule
A. Descriptor error ($z$) $\downarrow$ & 0.5350 \emph{[1.062]} & 0.7362 \emph{[1.189]} & 0.6803 \emph{[1.074]} & 0.7441 \emph{[1.139]} & 0.5137 \emph{[0.724]} & 0.5610 \emph{[0.545]} \\
B. Adherence $\uparrow$ & 0.6992 \emph{[0.718]} & 0.5429 \emph{[0.561]} & 0.5165 \emph{[0.505]} & 0.4925 \emph{[0.487]} & 0.5472 \emph{[0.178]} & 0.5012 \emph{[0.006]} \\
C. Spatial placement, lookup\nobreakdash-proof $\uparrow$ & 0.0101 \emph{[0.002]} & 0.0027 \emph{[-0.001]} & 0.0035 \emph{[-0.001]} & 0.0052 \emph{[-0.001]} & 0.0058 \emph{[-0.001]} & -0.0049 \emph{[0.005]} \\
D. Marginal realism $\downarrow$ & 0.2242 \emph{[0.247]} & 0.3661 \emph{[0.368]} & 0.1025 \emph{[0.076]} & 0.3484 \emph{[0.345]} & 0.0640 \emph{[0.145]} & 0.0268 \emph{[0.031]} \\
\bottomrule
\end{tabular}
}
  \caption{The four families at full context, with the corresponding
  random-context value shown in brackets.}
  \label{tab:generation}
\end{table}

Generation is scored on four families, each per clip and paired
(Figure~\ref{fig:generation}, Table~\ref{tab:generation}).
\textbf{A, descriptor error}: $z$-scored MAE between the local descriptor
recomputed from the sample and the true clip's, so $1.0$ is the error of an
unrelated clip. \textbf{B, adherence}: agreement with the descriptor the model
was \emph{given}, not the true one. \textbf{C, spatial placement}: map
correlation against the clip's own electrodes minus the same map scored against
a different clip of the same assay: the component a fixed site map cannot
reproduce. \textbf{D, marginal realism}: relative error on the canonical short-gap
rates.
Average precision has no place in this table: it ranks a probability map, so
the conditional mean maximizes it, and the site map of Section~\ref{sec:tasks}
duly outranks every learned arm on it (Figure~\ref{fig:tasks}).

We outperform the matched generative peer in all four families and every other
learned model in three; the U-Net leads on marginal realism.
Family C is the only one a site map cannot reproduce, and there the U-Net reaches
$0.0035$ against our $0.0101$. Context makes the U-Net \emph{worse} on the
family it wins, so that win is a pooled statistic and not a conditioning result
(Appendix~\ref{app:marginals}).

MaskGIT-flat gains more from conditioning: its median improvement is $0.2561$
on $848/1116$ clips, versus our $0.2108$ on $772/1116$
($q = 3.6\times10^{-37}$). \textbf{We therefore make no
claim that our conditioning is stronger.}

Our model instead shows stronger per-clip discrimination: its within-assay
correlation with the true ROI spike count is an order of magnitude above
MaskGIT-flat's, and it alone exceeds the arithmetic control in all three
settings. No arm combines discrimination with calibration; our decoded field
over-counts (Appendix~\ref{app:counts}).

\subsection{Design choices and their controls}
\label{sec:stages}

Every choice in Section~\ref{sec:method} has a control, tabulated in
Appendix~\ref{app:arch} and~\ref{app:nulls}. The blank route leaves all $32$
level-1 parents available for content, where a dense arm at the same seed and
schedule uses $25$ and shares $13$ with blanks (AP $0.0501$ against $0.0358$).
Every
ladder level contributes: re-decoding the same codes at cumulative depth adds
$+0.0832$ then $+0.1202$ AP on $100\%$ of clips. Patch size is a trade-off and
not an optimum (Table~\ref{tab:patchsweep}).

The two conditioning codes are complementary: the clip code better predicts
the temporal marginal ($R^2 = 0.9027$ versus $0.6063$), whereas the assay code
reaches $\Delta$NLL $0.3036$ against the clip code's $0.2072$; neither predicts
the temporal centroid. The texton basis sits on level~1 because
$z_1 > z_1{+}z_2 > \mathrm{flat}$ is monotone across all three arms, and the
motif prior's median rank is $9$ of $961$ against the strongest null's $53$.
Final adaptation raises MRR under emitted activity maps from $0.1898$ to
$0.2035$, against an oracle ceiling of $0.2149$, an effect $11.4$ times the
seed standard deviation. Adapting to hard maps instead gives $0.1427$, worse
than not adapting at all.

\section{Limitations}
\label{sec:limitations}

The corpus is biologically small: \NAssays\ assays from at most six organoids
and two patients, including three pairs with the same acquisition trace, and no
split here is cross-preparation. We therefore claim
\emph{scalability}, a parameter cost that stays flat in the number of
assays, and not generalization to unseen preparations. Removing the per-assay table is not the
absence of memorization, and Section~\ref{sec:reuse} shows the alphabet is
shared among training assays, not that it transfers
(Appendix~\ref{app:limits}).

Two comparison classes favor other methods. The static per-assay site map is
the strongest within-assay spatial reference because it stores assay-specific
structure (Section~\ref{sec:tasks}). The matched peer wins several pooled
summaries while we win the shape-sensitive terms, and the U-Net wins marginal
realism (Table~\ref{tab:marginals}). Because pooled statistics are less
sensitive to conditioning, our claims rely on paired per-clip tests. The
CVAE's latent collapses despite standard countermeasures, leaving the
continuous-latent question open (Appendix~\ref{app:baselines}).

Token granularity is the main weakness. A $(6,15,14)$ patch spans $210$
electrodes and $36$\,ms; the prior reaches $98\%$ of the site-level oracle-code
reference but only $5\%$ of the voxel reference, making within-token
\emph{timing} the principal failure mode. Two fixes failed
(Appendix~\ref{app:blur}), and smaller patches collapse the prior
(Appendix~\ref{app:limits}). The decoded field over-produces spikes; our per-clip
\emph{discrimination} is the strongest of any arm (Section~\ref{sec:generation}),
but the two are separate properties and no arm here has both. Finally, our tokenizer pairs a transformer core with a convolutional
stem and renderer while the matched peer's is fully convolutional, so this is
not a controlled comparison: that arm also changes the codebook structure and
the blank route (Appendix~\ref{app:limits}).

\section{Conclusion}
\label{sec:conclusion}

Discrete motifs are a workable representation for ultra-sparse HD-MEA spike
volumes. Learned once over a corpus spanning cultured organoid tissue and acute
\emph{ex vivo} human hippocampus, the alphabet achieves \CeilingFactor$\times$
the voxel-level reconstruction AP of a matched flat tokenizer given the true
codes, and is measurably reused across assays. A factorized prior over it
achieves $1.4$--$2.6\times$ the site-level AP of the matched peer, with no
learned per-assay parameter table, while within-token timing remains the
dominant failure mode and a static site map the stronger spatial predictor
(Section~\ref{sec:limitations}). Neither the blank route nor the where/what
factorization is specific to spike trains.

\section*{Ethics statement}

No new human or animal data were collected for this work. Both corpora are
open-access, de-identified secondary data obtained from the DANDI Archive
\citep{dandi000732, dandi001132}, and we report the approvals of the studies
that generated them.

The \emph{ex vivo} hippocampal recordings were obtained from tissue removed
during temporal lobectomy with hippocampectomy for drug-refractory epilepsy.
The tissue was resected for a clinically indicated reason and would otherwise
have been discarded. The source study reports that informed consent was obtained from
patients prior to surgical resection, that there was no participant
compensation, and that the work was approved by the University of California,
San Francisco institutional review board \citep{andrews2024hippocampal}.

The organoid recordings are from human iPSC-derived brain organoids, which do
not constitute human subjects research. The originating study reports that the iPSC line
was reprogrammed from skin fibroblasts obtained by biopsy following written
informed consent from the donor, under Washington University School of
Medicine institutional review board approvals 201104178 and 201306108
\citep{sharf2022organoids}. That study is the source of the organoid
recordings used here and provides the relevant consent information; the later
compilation that re-deposits them reports no human ethics approval of its own.

The archive records neither participant identifiers nor, for the organoid
recordings, the mapping from assay to preparation; we note in
Section~\ref{sec:data} that this limits what the corpus can support
scientifically, and it also means no re-identification risk is introduced by
releasing our per-assay provenance table.

On dual use: a generative forward model of spontaneous activity in cultured and
\emph{ex vivo} neural tissue is a research instrument for simulation,
null-model construction and closed-loop experiment design. We make no clinical
claim, the model is not diagnostic, and nothing here should be read as
characterizing the donors whose tissue produced the recordings.

\section*{Reproducibility statement}

Every number and figure panel is generated from a committed data artifact
rather than entered manually. Each reported value is therefore traceable to the
run that produced it. The  code release
(\url{https://github.com/tanveerderik/Organoid-Binary-Spike-Spatiotemporal-Data-Modeling})
 contains the
model, the training stages, all six evaluation arms including ours, the
evaluation harness, the
table and figure generators, and the commands that drive them. Its checkpoint
manifest records which checkpoint is shipped for each stage and how it was
selected.  
Evaluation is performed under a fixed protocol over an assay-balanced sample of
the test split (\NTestClips\ clips, nine per assay, across all \NAssays\
assays), with the task axis
averaging \NTaskMC\ Monte Carlo samples per clip under seed \NTaskSeed\ and
the distributional battery drawing one sample per rung. Pinning fixes the clip
set, the protocol and the committed artifacts; the battery's draws are
unseeded, so its statistics carry the finite-sample variation quantified in
Appendix~\ref{app:metrics} rather than reproducing bit for bit.
The clips are sampled from each assay's held-out windows with replacement, but
the draw is seeded, so every arm is scored on identical clips in identical order
(Appendix~\ref{app:metrics}). Data are the two
public dandisets cited above; Appendix~\ref{app:preproc} gives the
preprocessing chain in full, including which of the two stored spike
representations is read.

\section*{Use of AI statement}

Large language model assistance (Claude, Anthropic; ChatGPT, OpenAI) was used
throughout this project, largely in an agentic coding setting. Specifically, it
assisted with implementation of the model, training stages, baseline methods,
evaluation harness, analysis scripts, and figure and table generators. It was
also used for initial manuscript drafts and subsequent editing, code review and
debugging, methodological and experimental-design critique, literature search
and summarization, and preparation of this submission.

The research direction, the hypotheses, the experimental design, and every
claim made here are the authors'. No number in this paper was produced by a
language model: all results come from the committed artifacts of executed runs,
and the generator scripts exist precisely so that no value can be introduced by
transcription. The authors verified the results, read the code, and are
responsible for the paper's content, including any errors.

\section*{Funding}

M.A.M.-R. was supported by Schmidt Futures (SF857); the National Human Genome
Research Institute (RM1HG011543); the National Science Foundation (2515389);
the California Institute for Regenerative Medicine (DISC4-16285 and
DISC4-16337); the University of California Office of the President
(M25PR9045); the National Institute of Mental Health (U24MH132628); and the
National Institute of Neurological Disorders and Stroke (U24NS146314). The content is solely the responsibility of the authors and does
not necessarily represent the official views of the National Institutes of
Health, the National Science Foundation, CIRM, or any other agency of the State
of California. The work at Rensselaer Polytechnic Institute received no
specific funding; the grants listed above supported M.A.M.-R. and not this
study as a whole.

\section*{Competing interests}

M.A.M.-R. is a named inventor on patent applications relating to data
processing of high-throughput electrophysiology, and is an advisor for Atoll
Financial Group. The other authors declare no competing interests.

\bibliography{refs}
\bibliographystyle{iclr2027_conference}

\appendix
\setcounter{figure}{0}
\setcounter{table}{0}
\renewcommand{\thefigure}{S\arabic{figure}}
\renewcommand{\thetable}{S\arabic{table}}
\section{Data provenance}
\label{app:data}

Table~\ref{tab:provenance} lists every assay used, including what the archive
does and does not report. It is generated from the provenance artifact in the
code release. The recordings contain \CorpusMin\ minutes in total, or
\CorpusHours\ hours: \OrganoidMin\ minutes from organoid slices and \SliceMin\
minutes from \emph{ex vivo} hippocampal slices. Their durations vary
substantially, from \DurMinS\,s to \DurMaxS\,s. Spike extraction retains only
the burst windows, which contain \WindowedMin\ minutes of activity across
\NWindows\ windows, split into \TrainMin/\ValMin/\TestMin\ minutes for training,
validation and testing.

The biological cohort behind those recordings is smaller than the assay count, and the archive documents it only in part. Every file in the organoid dandiset names the same laboratory and the same experimenter, describes itself as intrinsic activity in a human brain organoid slice, and carries a MaxWell device string, so all \NOrganoidAssays\ of them are array recordings of sectioned organoids and none is a whole-organoid probe recording. The study behind that material reports six organoids recorded on such arrays, named L1--L6 and distinguished by array number in the accompanying data deposit, all grown from the single induced pluripotent stem cell line F12442.4 reprogrammed from skin fibroblasts of one consenting tissue donor \citep{sharf2022organoids}. We therefore describe the organoid material by organoids, cell line and donor, and the hippocampal material by slices and patients. The same source includes repeated recordings of one organoid across developmental time points, control and diazepam conditions, and four-hour interval series, so several assays can come from one organoid. A later study compiles this deposit together with recordings of its own into a larger cohort of eight human organoids \citep{vandermolen2026preconfigured}; that cohort is not the corpus used here, and we do not describe our assays by it.

The \NSliceAssays\ hippocampal assays come from two patients, a woman of $52$ and a man of $35$, and form two slice preparations: nine assays under the subject prefix \texttt{10F} and four under \texttt{11G}. The archive files each slice as its own subject, so its subject count is a slice count, and the source study reports a cohort of seven patients \citep{andrews2024hippocampal}.

\paragraph{Identity the conversion did not preserve.}
The organoid conversion wrote \texttt{neuroconv} defaults into every identity
field: one subject \texttt{sub-U}, one age range covering a human lifetime, one
session timestamp shared by all \NOrganoidAssays\ files, a UUID identifier, and a
session description saying only that the file was generated automatically. We
searched each file for the MaxWell array number, the original raw HDF5 file
name, and the experiment group. We searched all metadata fields and the raw
bytes of each complete file but found none of them. No assay in
Table~\ref{tab:provenance} can therefore be assigned to one of L1--L6, and none
can be assigned an experiment group or a drug condition. The
routing hash is the only resolution the files retain, and it gives
\NOrganoidRoutings\ distinct configurations for \NOrganoidAssays\ assays. A
shared routing configuration is evidence of a shared array, and therefore of a
shared organoid, but not of a shared recording: an organoid measured twice keeps
its routing.

\paragraph{Three pairs of assay files hold the same acquisition trace.}
Comparing the stored signal rather than the metadata shows what the routing hash
cannot. In three pairs the acquisition dataset is identical: \texttt{obj-1e5gdym}
with \texttt{obj-7c50zw}, \texttt{obj-1cs0t8m} with \texttt{obj-at6hmv}, and
\texttt{obj-1q4jnyp} with \texttt{obj-en3uzn}. Decompressed and compared sample
by sample over the whole array and the whole recording, each pair agrees on every
one of its $3.65$--$3.67\times10^{9}$ values. The two members of a pair are still
different files: they differ in size, in whole-file checksum, in the NWB version
they declare and in whether the electrode table carries redundant position
columns, so each pair is one recording converted twice, not one file
deposited twice. Nothing in the archive says why, and we do not infer a cause.

What the model consumes is not identical. The two conversions were spike-sorted
separately and sorting is not deterministic, so no burst window in one copy
carries the same spike event set as any window in the other. For a typical
window the best exact-event agreement with the other copy is a Jaccard index of
$0.002$--$0.004$, and the two copies differ by $3$--$18\%$ in total spike count
and by $6$--$26\%$ in active sites, with correspondingly different short-gap
rates and local descriptors. The windows do cover the same recording times, to a
median of $4$--$54$\,ms and never more than $0.34$\,s.

Because each copy is split in acquisition order, and the two orderings agree,
the same segment falls in the same split in both copies. Over the $95$ segments
the three pairs share, the split assignment differs for only one segment, which
is assigned to validation in one copy and test in the other. No segment is in
the training split of one copy and the test split of the other. We therefore
keep all \NAssays\ assays. The
duplication is a property of the archive rather than of the experiment, the
model was trained on \NAssays\ assays, and no result has been recomputed on a
reduced corpus.

\begin{table}[htbp]
  \centering
  \footnotesize
  \resizebox{\textwidth}{!}{\begin{tabular}{r l l l l l c r r l}
\toprule
idx & assay & dandiset & prep. & subject & age & sex & chan. & dur.\,(s) & routing \\
\midrule
0 & sub-U\_ses-20231119T113000\_obj-13gb78s & 000732 & organoid & U & P1D/P100Y & U & 1020 & 180 & \texttt{13207200a2} \\
1 & sub-U\_ses-20231119T113000\_obj-15bpqpa & 000732 & organoid & U & P1D/P100Y & U & 1018 & 862 & \texttt{7415ef1a7d} \\
2 & sub-U\_ses-20231119T113000\_obj-195c1dk & 000732 & organoid & U & P1D/P100Y & U & 1020 & 180 & \texttt{cead3bf0a8} \\
3 & sub-U\_ses-20231119T113000\_obj-1cs0t8m$\dagger$ & 000732 & organoid & U & P1D/P100Y & U & 1020 & 180 & \texttt{173ca3dd33} \\
4 & sub-U\_ses-20231119T113000\_obj-1e5gdym$\dagger$ & 000732 & organoid & U & P1D/P100Y & U & 1020 & 180 & \texttt{c005c7a3f4} \\
5 & sub-U\_ses-20231119T113000\_obj-1kcinv8 & 000732 & organoid & U & P1D/P100Y & U & 841 & 1940 & \texttt{262898a949} \\
6 & sub-U\_ses-20231119T113000\_obj-1q4jnyp$\dagger$ & 000732 & organoid & U & P1D/P100Y & U & 1014 & 180 & \texttt{2eb3b59799} \\
8 & sub-U\_ses-20231119T113000\_obj-1yigkds & 000732 & organoid & U & P1D/P100Y & U & 1020 & 180 & \texttt{dabafefa61} \\
9 & sub-U\_ses-20231119T113000\_obj-1ywxnj8 & 000732 & organoid & U & P1D/P100Y & U & 1020 & 378 & \texttt{d59ef839d8} \\
10 & sub-U\_ses-20231119T113000\_obj-7c50zw$\dagger$ & 000732 & organoid & U & P1D/P100Y & U & 1020 & 180 & \texttt{c005c7a3f4} \\
11 & sub-U\_ses-20231119T113000\_obj-at6hmv$\dagger$ & 000732 & organoid & U & P1D/P100Y & U & 1020 & 180 & \texttt{173ca3dd33} \\
12 & sub-U\_ses-20231119T113000\_obj-cjtijx & 000732 & organoid & U & P1D/P100Y & U & 1020 & 180 & \texttt{13207200a2} \\
13 & sub-U\_ses-20231119T113000\_obj-en3uzn$\dagger$ & 000732 & organoid & U & P1D/P100Y & U & 1014 & 180 & \texttt{2eb3b59799} \\
14 & sub-U\_ses-20231119T113000\_obj-fvpkkt & 000732 & organoid & U & P1D/P100Y & U & 1020 & 180 & \texttt{34dc02c543} \\
15 & sub-U\_ses-20231119T113000\_obj-hom2cn & 000732 & organoid & U & P1D/P100Y & U & 1020 & 180 & \texttt{a2d71f17ba} \\
17 & sub-U\_ses-20231119T113000\_obj-oqeor8 & 000732 & organoid & U & P1D/P100Y & U & 1020 & 180 & \texttt{a86ef3df2c} \\
19 & sub-U\_ses-20231119T113000\_obj-skejyf & 000732 & organoid & U & P1D/P100Y & U & 1020 & 180 & \texttt{34dc02c543} \\
20 & sub-U\_ses-20231119T113000\_obj-zvgjcc & 000732 & organoid & U & P1D/P100Y & U & 1020 & 180 & \texttt{bb7f365328} \\
21 & sub-10F-0\_ses-20240928T165427 & 001132 & slice & 10F-0 & P52Y & F & 887 & 313 & \texttt{51f17a2f10} \\
22 & sub-10F-1\_ses-20240928T165018 & 001132 & slice & 10F-1 & P52Y & F & 1009 & 127 & \texttt{c52a5b44c7} \\
23 & sub-10F-2\_ses-20240928T165008 & 001132 & slice & 10F-2 & P52Y & F & 1009 & 213 & \texttt{c52a5b44c7} \\
24 & sub-10F-3\_ses-20240928T165423 & 001132 & slice & 10F-3 & P52Y & F & 1009 & 200 & \texttt{c52a5b44c7} \\
25 & sub-10F-4\_ses-20240928T165229 & 001132 & slice & 10F-4 & P52Y & F & 1009 & 310 & \texttt{c52a5b44c7} \\
26 & sub-10F-5\_ses-20240928T165424 & 001132 & slice & 10F-5 & P52Y & F & 1009 & 320 & \texttt{c52a5b44c7} \\
27 & sub-10F-6\_ses-20240928T165427 & 001132 & slice & 10F-6 & P52Y & F & 1009 & 360 & \texttt{c52a5b44c7} \\
28 & sub-10F-7\_ses-20240928T165426 & 001132 & slice & 10F-7 & P52Y & F & 1009 & 390 & \texttt{c52a5b44c7} \\
29 & sub-10F-8\_ses-20240928T165425 & 001132 & slice & 10F-8 & P52Y & F & 1009 & 390 & \texttt{c52a5b44c7} \\
30 & sub-11G-0\_ses-20240928T192015 & 001132 & slice & 11G-0 & P35Y & M & 894 & 200 & \texttt{cc989abd68} \\
31 & sub-11G-1\_ses-20240928T165430 & 001132 & slice & 11G-1 & P35Y & M & 894 & 390 & \texttt{cc989abd68} \\
32 & sub-11G-2\_ses-20240928T165427 & 001132 & slice & 11G-2 & P35Y & M & 894 & 510 & \texttt{cc989abd68} \\
33 & sub-11G-3\_ses-20240928T165427 & 001132 & slice & 11G-3 & P35Y & M & 894 & 360 & \texttt{cc989abd68} \\
\bottomrule
\end{tabular}
}
  \caption{The \NAssays\ assays. \emph{prep.}\ is the preparation type
  reported by the source dandiset. \emph{routing} is a hash of the routed
  electrode set: assays sharing a hash were recorded through the identical
  channel selection. Note that all \texttt{sub-U} rows carry the same subject
  field, age and session timestamp. These are conversion defaults written by
  \texttt{neuroconv}, not measurements, and they collapse a cohort of at most
  six organoids into one apparent subject. $\dagger$ marks the six assays
  that form the three pairs holding the same acquisition trace.}
  \label{tab:provenance}
\end{table}

\paragraph{Assays present in the archive but not used.}
Three further \texttt{sub-U} recordings are present in dandiset 000732 and are
not in the table: \texttt{obj-1wcxx1y} ($2187$\,s), \texttt{obj-m6lpfz}
($1167$\,s) and \texttt{obj-paunmv} ($2044$\,s). All three are targeted
sub-region recordings for which the spike-extraction stage produced no burst
windows, so there is nothing to tokenize; they are excluded because the
extraction stage produced no files, not because of any selection we made. No
recording was dropped
on the basis of its activity, its statistics, or how any model performed on it.

\section{Vocabulary sharing}
\label{app:reuse}

\paragraph{The entropy and overlap statistics in full.}
Conditioning on assay identity takes the code entropy from $6.2770$ nats to
$5.7057$, a retained fraction of $0.9090$ under a Miller--Madow correction. The
correction biases against the shared-alphabet conclusion, since plug-in entropy
is downward-biased at these sample sizes. Raw, unrarefied overlap tracks
assay length at $r = +0.96$ and inverts the ordering between preparation
types, which is why every overlap reported in Section~\ref{sec:reuse} is at
matched sampling effort. Rarefied overlap sits slightly below the label-shuffle
null ($0.1087$ against $0.1243 \pm 0.0023$, $z = -6.69$), so a small
assay-specific component does exist.

\paragraph{Ladder depth, in figures.}
Re-decoding the same codes at cumulative depth, the third level adds $+0.1202$
AP on $100\%$ of clips ($p = 3.5\times10^{-12}$), whereas the summary-statistic
analysis attributes only $+0.019$ to it.

\begin{table}[htbp]
  \centering
  \begin{tabular}{l r}
\toprule
quantity & value \\
\midrule
alphabet entries in use & 846 of 961 \\
used by $\geq 5$ assays, rarefied & 255 \\
\midrule
$H(\mathrm{code})$ & 6.2770 nats \\
$H(\mathrm{code} \mid \mathrm{assay})$ & 5.7057 nats \\
\textbf{retained fraction} & \textbf{0.9090} \\
\midrule
rarefied overlap, within organoid & 0.1095 \\
rarefied overlap, within slice & 0.1232 \\
rarefied overlap, across preparation types & 0.1033 \\
\emph{null} label-shuffled & 0.1243 $\pm$ 0.0023 \\
\bottomrule
\end{tabular}

  \caption{Vocabulary sharing across the \NAssays\ assays, measured on the
  frozen tokenizer over the test split. The blank token is excluded: it is
  $91.7\%$ of all tokens and every assay emits it, so including it would
  drive every overlap statistic to $\approx 1.0$ while measuring nothing.
  Overlap is rarefied to a common token budget and compared with a
  label-shuffle null. Discussed in Section~\ref{sec:reuse}.}
  \label{tab:reuse}
\end{table}

Rarefaction removes the strong dependence of overlap on sampling effort.
Recordings contribute
between $117$ and $1756$ tokens and raw pairwise overlap tracks that sampling
effort at $r = +0.96$, so the unrarefied matrix reports a within-organoid mean
of $0.394$ against a within-slice mean of $0.221$, a preparation difference
that is entirely a difference in assay length, and one that reverses under
matched effort. The reported budget is the largest common budget available for
all \NAssays\ assays; the ordering is stable across every budget from $40$ to $117$, and
above $117$ the confound visibly returns as longer assays begin to
dominate.

\section{Ethics and data use}
\label{app:ethics}

The ethics statement before the references gives the consent and approval
language reported by each source study. Additional detail: dandiset
\texttt{000732} is the deposit we downloaded and is the version used here. It no
longer resolves in the archive, and the same organoid material is distributed by
the originating study through Dryad, with a README that names the six organoids
and their array numbers \citep{sharf2022organoids}. Dandiset \texttt{001603} is
the later compilation that incorporates that material alongside recordings of
its own \citep{vandermolen2026preconfigured}. Dandiset \texttt{001132} is
deposited at version \texttt{0.241130.1903}. Both dandisets we use are marked
\texttt{dandi:OpenAccess} in their archive metadata. Neither carries participant identifiers. We collected no new
human or animal data, ran no experiment on tissue, and had no contact with
participants.

\section{Preprocessing}
\label{app:preproc}

\paragraph{Spike extraction.}
Each assay is sorted with SpyKING Circus~2 through SpikeInterface
\citep{buccino2020spikeinterface}: a fourth-order Butterworth bandpass at
$300$--$6000$\,Hz, a detection neighborhood of $100\,\mu\mathrm{m}$, and a
waveform window of $2.0$\,ms before and after the peak. Each curated unit is
written to its peak electrode.

\paragraph{Burst-window detection.}
Continuous recordings are reduced to burst windows before anything else runs.
The per-channel binary raster is summed across channels into a population
trace, smoothed with a $2000$-sample boxcar ($100$\,ms at $20$\,kHz, edge
padded), and its Hilbert envelope is normalized. The dominant frequency of the
mean-removed envelope below $5$\,Hz sets a characteristic burst period, and
peaks are taken from the smoothed trace with a minimum separation of $0.75$ of
that period and a prominence of $20$ spikes. Each accepted peak yields the
window $[t - 0.2\,\mathrm{s},\, t + 0.4\,\mathrm{s}]$, i.e.\ $600$\,ms with the
peak one third of the way in; windows extending past either end of the
recording are dropped. This produces the $2{,}133$ windows used here. The
extraction runs upstream of this code release, which begins at the stored
window arrays; the parameters above are the ones the extraction used.

\paragraph{Choice of stored spike representation.}
The extraction step writes two arrays per burst window, and reading the wrong
one changes every number in this paper. \texttt{binary\_unit\_burst} stores
\emph{point events}: one sample per spike at the unit's peak electrode, with a
mean run length of exactly $1.0$ samples. \texttt{binary\_ch\_burst} stores the
sorter's \emph{waveform extent}: the $4$\,ms window around each detection
painted onto every electrode in the neighborhood, with a mean run length of
$86.8$ samples. The loader reads the point-event family. Reading the waveform family instead inflates the
voxel rate by roughly $22\times$, and the regression test asserts both run
lengths so the two cannot be silently swapped.

\paragraph{Binning, cropping, padding.}
A stored window is $(120, 220, 12000)$ at one $20$\,kHz sample per bin, i.e.\
$50\,\mu$s per bin and $600$\,ms per window; there are $2{,}133$ such windows
across the \NAssays\ assays. Max-pooling by $120$ raw bins gives $6$\,ms
frames, so a frame records \emph{whether} an electrode fired in that interval
and not how many times. At the observed firing rate, this temporal pooling
discards little information, but it
is the reason a voxel is binary and not a count. A window yields $100$
pooled frames; a contiguous span of $50$ pooled frames (a $6000$-sample crop)
is drawn at random, then end-cropped to \ClipFrames\ frames, the nearest
multiple of the patch's temporal extent of $6$. The
array footprint is $120\times220$ and is padded on the width axis to
$120\times224$, a multiple of the patch's spatial extent of $14$. The result is
a $\ClipFrames\times120\times224$ binary volume of \ClipVoxels\ voxels.

\paragraph{The two occupancy rates.}
A whole $600$\,ms burst window has occupancy $7.59\times10^{-5}$. A clip is a
$288$\,ms span drawn inside such a window, and the draw lands preferentially on
the active part, so clip occupancy is about twice that: \VoxelRate, or
\SpikesPerClip\ spikes per clip. Every number in this paper is computed on
clips and quotes the clip rate. Both rates are measured on $48$ validation
clips at the shipped clip geometry, so they describe the data the model is fitted
to rather than a held-out estimate; they are descriptive statistics of the corpus and no result depends on them.

\section{Architecture and design curves}
\label{app:arch}

\paragraph{Level warm-up by loss weight.}
Levels are activated one at a time by ramping their loss weights. Scaling the
level outputs instead would rescale the vectors the EMA accumulates, biasing the
very target the codebook is converging to; staged activation by weight gave a
large improvement over activating all three simultaneously.

\paragraph{Deduplication criterion.}
The merge tests whether two entries are the same vector. Usage frequency
measures a different quantity, how often an entry is chosen, so a frequency
filter deletes rare but distinct motifs and leaves genuine duplicates in place.
An earlier version of this step used a frequency filter.

\begin{figure}[htbp]
  \centering
  \includegraphics[width=\textwidth]{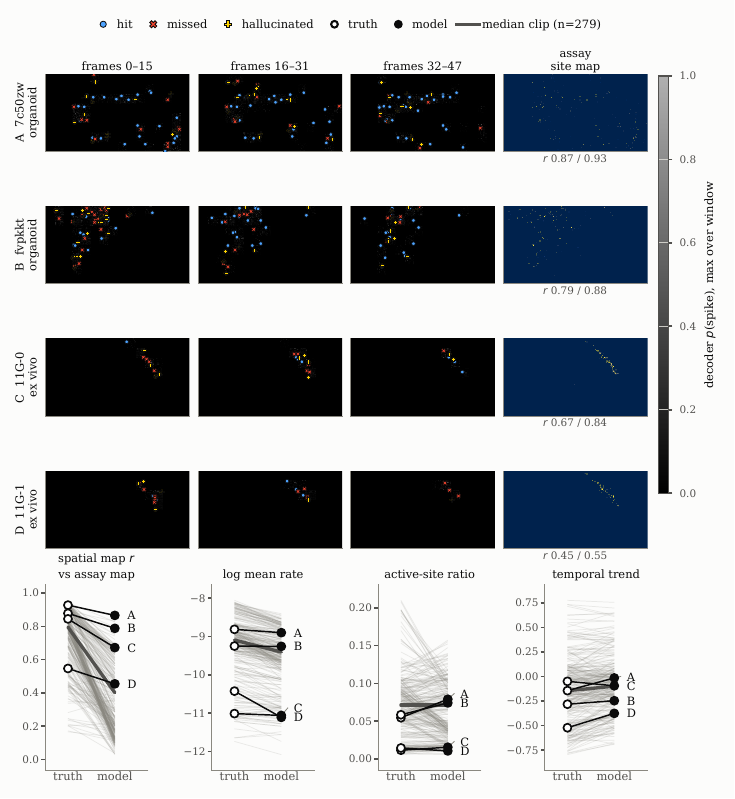}
  \caption{Tokenizer reconstruction, four assays, two per corpus. Marks are
  the thresholded reconstruction against the truth; the grayscale ground is the
  decoder's probability. The right column is the assay's training site map,
  the reference the correlation beneath each row is measured against. These
  rows attain $0.9335$, $0.8973$, $0.7962$ and $0.8315$ of their achievable map
  correlation and are therefore far from typical: over all $279$ test clips the
  median falls from $0.7955$ to $0.2335$, and only a fraction $0.0179$ of clips
  place better than their own truth does. The strip is paired (one segment per clip, heavy line the median paired change) because the pooled version
  disagrees: active-site ratio has marginal medians $0.0713$ and $0.0479$, a
  third apart, while the median \emph{per-clip} change is $0.0$ with a fraction
  $0.491$ of clips moving up. The rate deficit is the real weakness on display,
  a median $-0.3129$ in log mean rate with only a fraction $0.0538$ of clips
  over-producing. The tokenizer is given the true local code, so preserving it
  here is close to circular; Figure~\ref{fig:qual_gen} is where conditioning is
  actually tested.}
  \label{fig:qual_recon}
\end{figure}

Reconstructions from four assays, with their spatial-map and local-code
adherence measured on the assay-balanced evaluation sample, are in
Figure~\ref{fig:qual_recon}.

\paragraph{Stage controls, in figures.}
Against the strongest rung of the null ladder, the motif prior's cross-entropy
is lower by $2.04$ nats. The two conditioning codes are complementary: the
local code reaches $R^2$ $0.9027$ on the temporal marginal against the global
code's $0.6063$, while the global code reaches $\Delta$NLL $0.3036$ against the
local code's $0.2072$. No conditioning predicts the temporal centroid
($0.0224$--$0.1204$). A sparse encoder keeps all $32$ parent codes available for
content against $25$ for a dense arm at identical seed and schedule.

\paragraph{Architecture.}
A $3\times3\times3$ convolutional stem takes the single input channel to $8$ and
feeds a patch embedding; the encoder is then a $2$-layer transformer of width
$64$ with $4$ heads over the $8\times8\times16$ token grid, with global
attention. The decoder is its mirror with causal-in-time attention, followed by
a transposed-convolution patch renderer that maps a token embedding to its
$1260$ output voxels. The pipeline is therefore a convolution--transformer
hybrid. As noted in Section~\ref{sec:limitations}, the missing comparison is
between the transformer \emph{core} and a convolutional core, not between the
entire pipeline and a convolutional pipeline. The code
dimension is $64$. The residual ladder has $32$, $8$ and $4$ entries at levels
$1$, $2$ and $3$, each level a dense tree conditioned on the path above it, so
the tree has $32\times8\times4 = 1024$ leaves. Codebooks are EMA buffers and carry no gradient. Both MaskGIT priors are width $128$, $4$
layers, $4$ heads, dropout $0.1$; the activity prior additionally carries a
count head over $K_{\max} = 312$ bins and reads the ROI occupancy of a
$2\times2\times4$ partition of the grid into $16$ regions of $64$ cells, so the
shape of the hole enters as context at region granularity. There is no regional
count target; the count head predicts one total for the clip.

\paragraph{Deduplication.}
Merging is by pairwise relative distance at $0.05$, giving $961$ distinct
entries from $1024$ leaves. The result is insensitive to moderate changes in
the threshold: raising it from
$0.02$ to $0.2$ moves the number of colliding pairs only from $107$ to $121$,
and the median nearest-neighbor relative distance across the alphabet is
$0.543$. The merged group is a well-separated tail.

\paragraph{Patch size.}
\begin{table}[htbp]
  \centering
  \small
  \begin{tabular}{l r r r r r r}
\toprule
patch $(T,H,W)$ & voxels & tokens & blank \% & active tok. & spikes/active & capture$_{K=32}$ \\
\midrule
$(24,15,14)$ & 5040 & 256 & 79.6 & 52.2 & 4.01 & 0.0600 \\
$(12,30,14)$ & 5040 & 256 & 76.0 & 61.4 & 3.41 & 0.0617 \\
$(6,30,28)$ & 5040 & 256 & 73.1 & 68.9 & 3.04 & 0.0699 \\
$(12,15,14)$ & 2520 & 512 & 84.5 & 79.1 & 2.65 & 0.0607 \\
$(6,30,14)$ & 2520 & 512 & 81.8 & 92.9 & 2.25 & 0.0675 \\
$(24,15,7)$ & 2520 & 512 & 86.4 & 69.4 & 3.01 & 0.0559 \\
\textbf{$(6,15,14)$} (shipped) & 1260 & 1024 & 88.9 & 113.6 & 1.84 & 0.0866 \\
$(3,15,14)$ & 630 & 2048 & 92.7 & 148.9 & 1.41 & 0.1385 \\
$(6,15,7)$ & 630 & 2048 & 93.4 & 134.4 & 1.56 & 0.1060 \\
$(6,10,7)$ & 420 & 3072 & 95.3 & 143.3 & 1.46 & 0.1259 \\
$(3,15,7)$ & 315 & 4096 & 95.9 & 166.8 & 1.26 & 0.1809 \\
$(3,10,7)$ & 210 & 6144 & 97.2 & 173.9 & 1.20 & 0.2281 \\
$(2,8,7)$ & 112 & 11520 & 98.4 & 189.4 & 1.11 & 0.3799 \\
\bottomrule
\end{tabular}

  \caption{The patch-size sweep, at fixed data. \emph{capture}$_{K=32}$ is the
  fraction of patch variance a $32$-entry $k$-means alphabet recovers, i.e.\ a
  model-free proxy for how well a small alphabet can describe patches of that
  size. It improves monotonically as the patch shrinks, and so does the blank
  fraction: at $(3,15,14)$ the grid is $2048$ tokens of which $92.7\%$ are
  empty. The shipped choice is the largest patch at which capture is still
  rising steeply.}
  \label{tab:patchsweep}
\end{table}

The sweep is why we describe patch size as a design choice and not an
optimum. Reconstruction alone would favor the smallest patch. However, smaller
patches enlarge the token grid and increase the blank fraction; a prior trained
on a grid that is $97\%$ blank collapses toward blank predictions. The selected
patch size therefore reflects a tokenizer--prior trade-off whose limitations
are described in Section~\ref{sec:limitations}.

\paragraph{Ladder depth.}
Table~\ref{tab:ladderdepth} quantifies the contribution of each level.

\begin{table}[htbp]
  \centering
  \small
  \begin{tabular}{l r l}
\toprule
decode depth & AP $\uparrow$ & gain over previous \\
\midrule
$z_1$ & 0.0772 & -- \\
$z_1{+}z_2$ & 0.1604 & \textbf{$+$0.0832} (100\% of clips) \\
$z_1{+}z_2{+}z_3$ & 0.2806 & \textbf{$+$0.1202} (100\% of clips) \\
\bottomrule
\end{tabular}

  \caption{Decoding the \emph{same} codes at cumulative depth. Nothing is
  retrained between rows, so the only variable is how much of the residual sum
  reaches the decoder. The third level is not inert: it adds $+0.1202$ AP on
  $100\%$ of $64$ clips, even though a summary-statistic $\eta^2$ analysis of
  the same codes attributes only $+0.019$ to it. The effect has to be
  evaluated at the decoder output, not through a summary statistic.}
  \label{tab:ladderdepth}
\end{table}

\paragraph{Decoder-side conditioning.}
An earlier tokenizer carried a gated cross-attention branch giving the decoder
direct access to the conditioning codes, and it never became active: over $300$
epochs, $21$ paired conditional and unconditional validation series (AP, tolerant AP, best-$F_1$ and its threshold, the BCE and VQ losses, and the three reference reconstruction losses) are bit-identical, maximum absolute
difference exactly $0.0$. Because the reconstruction objective provides no
incentive to use this conditioning path, increasing the gate magnitude did not
activate it. In the shipped model, conditioning therefore reaches generation
through the priors and affects the tokenizer only through output constraints.

\paragraph{The residual after three levels.}
An adapter predicting a continuous mixture weight over the codebook hull, in
place of a discrete third level, logged $24$ diagnostic series identically zero
for $300$ epochs. What remains after the ladder is close to quantization noise
and about $4\%$ predictable in any basis we tried, so a fourth continuous stage
has nothing to fit.

\paragraph{Sparse encoder.}
Against a dense arm at identical seed, data and schedule, the sparse encoder
keeps all $32$ level-1 parent codes on content patches, while the dense arm
uses $25$ and shares $13$ of those $25$ with blank patches: more than half its alphabet is doing double duty. Validation AP is $0.0501$ against $0.0358$.
The dense arm ablates the sparse design as a \emph{package} and does not
isolate quantizer routing: two auxiliary losses key off the blank mask and go
inactive when nothing is blank. It answers ``why this encoder design'', not
``why this quantizer''.

\section{Training protocol}
\label{app:training}

Stages are trained in sequence and earlier stages are frozen thereafter; no
stage is trained jointly with another. The tokenizer runs $300$ epochs with the
three ladder levels activated in turn by ramping their loss weights, never by scaling the level outputs. A level scale $\neq 1$ rescales the vectors the
EMA accumulates and biases the target the codebook converges to. Staged
activation measured $3.4\times$ better than activating all three at once.

The local-context mapper trains for $300$ epochs against a multinomial
likelihood, with its marginal baseline accumulated online as a Dirichlet$(1)$
posterior mean across every batch of every epoch, not from a single
collection pass. The prior stages use warmup followed by cosine decay. The
selection metric differs across stages by design: the motif prior selects on MRR, the
activity prior on validation NLL, and the adaptation stage on validation MRR.
The activity prior in particular must \emph{not} select on a rank metric: rank metrics are blind to calibration, and the activity prior's output is a
probability that is later thresholded. Neither the activity prior's count
teacher nor its early stopping is allowed to act before the count handoff
completes, since a teacher-forced model is solving an easier problem and its
score must not set the selection bar.

\section{Hyperparameters}
\label{app:hparams}

Table~\ref{tab:hparams} is the full training configuration, one row per stage.

\begin{table}[htbp]
  \centering
  \scriptsize
  \resizebox{\textwidth}{!}{%
\begin{tabular}{l l r r l r r r r l}
\toprule
stage & what & lr & wd & schedule & warm & ep. & clip & pat. & selected on \\
\midrule
1 & Global-code mapper pretrain & $1\times 10^{-3}$ & $1\times 10^{-4}$ & none & -- & 200 & -- & 20 & spatial-map loss \\
2A & Tokenizer (3-level residual VQ-VAE) & $1\times 10^{-3}$ & $1\times 10^{-4}$ & cosine to $1\times 10^{-5}$ & -- & 300 & -- & 40 & val exact AUPRC \\
3 & Local-code mapper (texton multinomial) & $1\times 10^{-3}$ & $1\times 10^{-2}$ & cosine to $1\times 10^{-5}$ & -- & 300 & -- & 40 & val multinomial NLL \\
4A & Motif prior (MaskGIT over V=961) & $3\times 10^{-4}$ & $1\times 10^{-2}$ & warmup + cosine & 10 & 600 & 1.0 & 150 & val MRR \\
4B & Activity prior (where, plus count head) & $2\times 10^{-4}$ & $1\times 10^{-2}$ & warmup + cosine & 5 & 120 & 1.0 & 30 & val NLL \\
4C & Adaptation of the motif prior to emitted maps & $3\times 10^{-5}$ & $1\times 10^{-2}$ & warmup + cosine & 5 & 120 & 1.0 & 40 & val MRR \\
\bottomrule
\end{tabular}}

\vspace{0.6em}

\begin{minipage}{\textwidth}\footnotesize
Shared across every stage: batch size 4 with 8 accumulation steps, i.e.\ an effective batch of 32; CUDA autocast, FP16; dropout 0.1 in both priors; seed 0. Stages run in sequence and every earlier stage is frozen thereafter.\\[0.3em]
Per-stage settings. \textbf{1} $\lambda_{\mathrm{sep}}$ 0.0001; \textbf{2A} $\lambda_{\mathrm{ctx}}$ 2.8, $\lambda_{\mathrm{ctx\,field}}$ 0.05, $\lambda_{\mathrm{code\,norm}}$ 0.1, pos weight 100 to 1 over 100 epochs; \textbf{3} n textons 128, texton basis z1, batches per epoch 120; \textbf{4A} $\lambda_{\mathrm{z1\,neighbor\,ce}}$ 0.05, $\lambda_{\mathrm{ctx}}$ 0.25, $\lambda_{\mathrm{adj}}$ 0.25, $\lambda_{\mathrm{spatial}}$ 0.25, full mask prob 0.15; \textbf{4B} $\lambda_{\mathrm{bce}}$ 1.0, pos weight 1.0, $\lambda_{\mathrm{count}}$ 0.065, $\lambda_{\mathrm{spatial}}$ 0.02, random mask prob 0.5; \textbf{4C} emitted-map ramp 0 to 0.8 over 20 epochs, readout gumbel, soft field on.
\end{minipage}

  \caption{Every setting the shipped pipeline trains under, one row per stage.
The values are read out of the training source automatically and not
transcribed, so a renamed argument raises instead of leaving a stale
value in the table. Only the shipped stages appear.}
  \label{tab:hparams}
\end{table}

Two settings differ from the others by design. The activity prior is the only
stage selected on a likelihood: rank metrics are blind to calibration, and its
output is a probability that is later thresholded. The adaptation stage runs an
order of magnitude below the stage it adapts, since it corrects a mismatch
between two trained modules instead of learning the task again.

\section{Loss terms}
\label{app:objectives}

Section~\ref{sec:method} names each objective by symbol.
Table~\ref{tab:objectives} is where each one is defined.

\begin{table}[htbp]
  \centering
  \scriptsize
  \begin{tabular}{@{}>{\raggedright\arraybackslash}p{0.24\textwidth} l >{\raggedright\arraybackslash}p{0.44\textwidth}@{}}
\toprule
term & coefficient & what it constrains \\
\midrule
\multicolumn{3}{l}{\textbf{Stage 1}\quad $\mathcal{L}_{\mathrm{global}}$} \\[0.15em]
\quad support, token grid & outside 2.0, mass 1.0 & cover the recording's active sites, penalize mass outside them, match the total \\
\quad support, full array & outside 3.0, mass 2.0 & the same at electrode resolution; ramped in as the token arm decays \\
\quad separation & 0.0001 & repel the predicted maps of recordings whose true maps already differ, at margin 0.25 \\
\quad adjacency & 0.5 & match the recording's same-site short-gap rates \\
\addlinespace
\multicolumn{3}{l}{\textbf{Stage 2A}\quad $\mathcal{L}_{\mathrm{tok}}$} \\[0.15em]
\quad tolerant spike reconstruction & 100 to 1 over 100 epochs & exact BCE at a decaying positive weight, plus max-pooled hit, peak-margin and multi-count terms \\
\quad vector quantization & 1.0 & commitment at $\beta = 0.25$ plus a usage-entropy term at 0.001 against a detached codebook; codebooks are EMA at decay 0.95 \\
\quad context & 2.8 & the nine local moments recomputed from the decoder's own logits must match the clip's true descriptor \\
\quad context field & 0.05 & the same nine moments per patch rather than per clip \\
\quad spatial, token grid & 0.001 & no mass where the recording's support map forbids it \\
\quad spatial, full array & 0.0001 & the same at electrode resolution, ramped in late \\
\quad short-gap rate & 0.01 & same-site inter-spike gaps must not exceed or fall short of the recording's measured rates \\
\quad code norm & 0.1 & a soft ceiling on encoder output norms, so codes do not drift out of the EMA's reach \\
\quad blank hinge & 0.05 & empty patches must not produce spike logits \\
\quad blank/active separation & 0.01 & the decoder's blank and content embeddings stay apart \\
\quad encoder isotropy & 1.0 & a variance floor and an off-diagonal correlation penalty, against encoder collapse \\
\addlinespace
\multicolumn{3}{l}{\textbf{Stage 3}\quad $\mathcal{L}_{\mathrm{local}}$} \\[0.15em]
\quad texton multinomial NLL & 1.0 & the predicted distribution over the 128-texton basis must match the clip's soft texton histogram \\
\quad flat-code multinomial NLL & 1.0 & the same against the $V{=}961$ alphabet \\
\quad regional summary & 1.0 & squared error on per-code mean time and per-frame active fraction \\
\addlinespace
\multicolumn{3}{l}{\textbf{Stage 4A / 4C}\quad $\mathcal{L}_{M}$} \\[0.15em]
\quad categorical cross-entropy & 1.0 & the correct motif at each masked active cell \\
\quad neighborhood CE & 0.05 & partial credit over the five codebook-nearest alternatives at temperature 0.25, so a near miss in alphabet geometry is not a total miss \\
\quad expected code distance & 0.05 & the whole predicted distribution is pulled toward the target's neighborhood, not just its mode \\
\quad context & 0.25 & soft-decode the motif logits through the frozen tokenizer; the result's nine moments must match the clip \\
\quad context field & 0.05 & the same moment match evaluated per patch, so a clip cannot be right on average and wrong everywhere \\
\quad adjacency & 0.25 & the decoded volume's short-gap rates must match the recording's \\
\quad spatial & 0.25 & no decoded mass outside the recording's support map \\
\addlinespace
\multicolumn{3}{l}{\textbf{Stage 4B}\quad $\mathcal{L}_{A}$} \\[0.15em]
\quad per-cell BCE & 1.0 & active against blank at each token cell, at positive weight 1.0 -- deliberately not the class-balancing ratio, which over-produces \\
\quad count & 0.065 & one categorical over the clip's total active count, with an ordinal neighbor term at 0.25 and an expected-distance term at 0.05 so a near-miss count is scored as one \\
\quad temporal co-activation & 0.1 & the predicted field's co-activation rate along time must match the batch's \\
\quad spatial co-activation & 0.1 & the same co-activation match taken across the array instead of along time \\
\quad spatial support & 0.02 & no predicted activity in columns the true map leaves empty \\
\bottomrule
\end{tabular}

\vspace{0.4em}

\begin{minipage}{\textwidth}\footnotesize
Coefficients are constant unless noted; the ramps and curricula are in Appendix~\ref{app:training}. Present in the code and inactive in the shipped configuration: Stage 2A, token-profile entropy (the within-token blur experiment; reported as a failure and not shipped); Stage 4A/4C, the second entry of loss\_weights (0.1) (the alpha head it weighted was removed; the motif prior reads loss\_weights[0] only); Stage 2A, false-positive weighting (tolerant\_spike\_loss is called with fp\_weight 0, so the hallucination penalty is inactive); Stage 2A, latent masking, CFG dropout (both off in the shipped run).
\end{minipage}

  \caption{Every term of every stage's objective, with its coefficient and
  what it constrains. Section~\ref{sec:method} names these terms; this is where
  they are defined. Coefficients are read out of the source automatically and
  not transcribed. Two coefficients require clarification because their
  effective run-time values differ from the first values visible in the source:
  Stage 2A's context weight is rebound
  at run time from $0.1$ to $2.8$, and Stage 4B's coefficients come from the
  second of two configuration dictionaries, the first of which the loss never
  reads.}
  \label{tab:objectives}
\end{table}

\section{Metric definitions and multiple comparisons}
\label{app:metrics}

\paragraph{Hidden fraction per evaluation setting.}
Free generation hides $1.000$ of the token grid, causal $0.587$, noncausal
$0.396$ and spatial $0.339$. These are fractions of the \emph{token} grid, not
of the frame axis: a causal prefix keeps $25$--$75\%$ of frames and a noncausal
hole covers a contiguous $30\%$ of them, but holes are snapped to the patch
lattice before scoring, which rounds a partly covered token into the hole.
Because the base rate differs by column, a number is comparable
across arms within a column and never across columns; no aggregate over the four
settings is reported anywhere in this paper.

\paragraph{Average precision.}
Every AP value is computed stepwise, never trapezoidally:
linear interpolation in
PR space is invalid \citep{davis2006prcurves} and here inflates the saturating
flat tokenizer by $+0.2239$ against our $+0.0015$. \emph{Site-level} AP scores
whether the right electrodes are active anywhere in the clip; \emph{voxel-level}
AP scores whether the right electrode is active in the right frame. At this
sparsity the two answer different questions and are never averaged together.
The canvas is padded from $220$ to $224$ columns so the width divides the patch
grid, and those four columns are inside the scored region: they hold no
electrode, so they are negatives for every arm alike and neither adds nor
removes a positive. The descriptor and spatial-map metrics un-pad first, since
the stored per-assay map is unpadded.

\paragraph{Distributional metrics.}
Relative error terms are $|{\hat{s} - s}| / s$ on a named statistic computed per clip. The statistics are the spike rate, the $k$-frame persistence rates, the mean avalanche size, the burst rate and the mean inter-spike interval. KS terms are two-sample Kolmogorov--Smirnov distances on the
inter-spike-interval and avalanche-size distributions. Bin edges and feature lists come from one shared definition that every arm imports, so no two arms can be scored on different bins. Two statistics are excluded as degenerate
before any test is run: spatial co-activation is degenerate on $99.2\%$ of clips and single-frame persistence on $28.9\%$. The exclusion is applied
identically to every arm.

\paragraph{Against a single generation composite.}
Selecting the activity prior on a single scalar combining them fails a basic
check: placing the \emph{oracle} activity map into that composite \emph{loses}
by $0.0127$. A selection metric an oracle cannot win is not measuring what it
claims to, so the activity prior selects on validation NLL and the four
families are reported separately.

\paragraph{Multiple comparisons.}
All paired tests are Wilcoxon signed-rank, on the same clips in the same order,
with pairing asserted, not assumed. Benjamini--Hochberg
\citep{benjamini1995fdr} is applied jointly over the $32$ task-axis comparisons (two metric families by four settings by every learned arm against ours), and $q$ is reported for that family. Analyses outside it are
identified where they appear and report their own statistic: the ladder-depth
increment is a paired test within one model and quotes $p$, and the
vocabulary-overlap null is a $z$ score against a label shuffle.

\paragraph{Scored clips.}
The temporal test split holds \NTest\ burst windows, unevenly distributed over
the \NAssays\ assays. Evaluation does not walk that split directly. It scores an
assay-balanced set of \NTestClips\ clips, nine per assay, each drawn from that
assay's held-out windows with replacement and cropped under a fixed seed, so a
long recording does not dominate and a short one is still represented. The draw
is deterministic: the same seed yields the same \NTestClips\ clips in the same
order for every arm, which is what makes the comparisons paired. Nine per assay
is the test quota; training and validation use thirty and six.

The four settings are sampled per clip, so a setting scores a subset of those
clips and not all of them: $279$ on free generation, $278$ on causal and on
noncausal, and $243$ on spatial. A clip is dropped from a setting when its hole
contains no spike, since average precision is undefined with no positive, and a
spatial box at this occupancy encloses no spike more often than a temporal hole
does. The task axis
additionally averages \NTaskMC\ Monte Carlo samples per clip under seed
\NTaskSeed; the distributional battery draws one sample per clip per rung and is
not seeded, so repeating it redraws those samples.

The clip set is drawn before evaluation and cannot have been chosen after
seeing a result: the draw is seeded, and the loader then reads its
\NTestClips\ indices unshuffled, so the order is fixed and the harness seed
governs only the model's own randomness, namely Monte Carlo draws, the Gumbel
readout and mask draws. Every arm therefore sees
identical clips in an identical order, which is the precondition for the paired tests. The comparison harness verifies it instead of assuming it, by requiring two runs to agree bitwise on a model-free arm computed from the same
ground truth before it will pair them.

The cost sets the budget. One clip in the task axis requires an
iterative MaskGIT decode per Monte Carlo sample, in four settings, for each of
six arms and their nulls; the full protocol reported here is several GPU-hours.
An earlier version of this work ran $12$ and $8$ batches, which, because the prefix is ordered by assay, reached only $6$ and $4$ assays. Those
numbers are not reported here. The distribution terms in particular are
finite-sample statistics whose \emph{absolute} value moves with the budget: one
composite we measured shifted by $+0.036$ between $4$ and $8$ validation batches
on a fixed checkpoint, which is why no number is carried between protocols.

\section{Full conditioning ladders}
\label{app:ladders}

Section~\ref{sec:generation} reports each family at two rungs, the arm's own
random-context control and full context. Table~\ref{tab:fullladders} gives all
five, which is what shows whether an arm improves \emph{monotonically} as
context is added or only differs at the endpoints. Among the families where
improvement is the right expectation, A, C and D, two arms fail and the
endpoint view hides it: the U-Net degrades on D as context accumulates, and the
GLM is flat across every rung. Family B is excluded from that reading because
it scores obedience to the descriptor supplied, not proximity to the
true clip (Table~\ref{tab:fullladders}). \emph{local only} is a
control and not a rung, since it withholds the assay code entirely.

\begin{table}[htbp]
  \centering
  \footnotesize
  \begin{tabular}{l r r r r r}
\toprule
arm & random & local only & global only & global partial local & global full local \\
\midrule
\multicolumn{6}{l}{\textbf{A. Descriptor error ($z$)} $\downarrow$} \\
\quad Ours (4C+soft) & 1.0623 & 0.8473 & 0.6738 & 0.6172 & 0.5350 \\
\quad MaskGIT-flat & 1.1892 & 0.9595 & 0.8347 & 0.7528 & 0.7362 \\
\quad 3D U-Net (det.)$^\dagger$ & 1.0741 & 0.8415 & 0.7711 & 0.6740 & 0.6803 \\
\quad 3D CVAE & 1.1388 & 0.9099 & 0.8373 & 0.7581 & 0.7441 \\
\quad Dich. Gaussian & 0.7243 & 0.5183 & 0.6009 & 0.5298 & 0.5137 \\
\quad Coupled GLM & 0.5447 & 0.5584 & 0.5560 & 0.5648 & 0.5610 \\
\midrule
\multicolumn{6}{l}{\textbf{B. Adherence} $\uparrow$} \\
\quad Ours (4C+soft) & 0.7183 & 0.3081 & 0.6945 & 0.7212 & 0.6992 \\
\quad MaskGIT-flat & 0.5608 & 0.2535 & 0.5457 & 0.4971 & 0.5429 \\
\quad 3D U-Net (det.)$^\dagger$ & 0.5047 & 0.3280 & 0.5042 & 0.5120 & 0.5165 \\
\quad 3D CVAE & 0.4872 & 0.3364 & 0.4885 & 0.4751 & 0.4925 \\
\quad Dich. Gaussian & 0.1779 & 0.5150 & 0.5210 & 0.5416 & 0.5472 \\
\quad Coupled GLM & 0.0057 & 0.4973 & 0.5094 & 0.5039 & 0.5012 \\
\midrule
\multicolumn{6}{l}{\textbf{C. Spatial placement, lookup\nobreakdash-proof} $\uparrow$} \\
\quad Ours (4C+soft) & 0.0018 & -0.0012 & -0.0058 & 0.0104 & 0.0101 \\
\quad MaskGIT-flat & -0.0008 & 0.0003 & -0.0040 & 0.0039 & 0.0027 \\
\quad 3D U-Net (det.)$^\dagger$ & -0.0014 & -0.0004 & 0.0010 & 0.0047 & 0.0035 \\
\quad 3D CVAE & -0.0009 & 0.0010 & -0.0014 & -0.0003 & 0.0052 \\
\quad Dich. Gaussian & -0.0009 & 0.0049 & 0.0018 & 0.0014 & 0.0058 \\
\quad Coupled GLM & 0.0048 & -0.0052 & 0.0066 & -0.0028 & -0.0049 \\
\midrule
\multicolumn{6}{l}{\textbf{D. Marginal realism} $\downarrow$} \\
\quad Ours (4C+soft) & 0.2467 & 0.2285 & 0.2523 & 0.2145 & 0.2242 \\
\quad MaskGIT-flat & 0.3684 & 0.3808 & 0.3704 & 0.3842 & 0.3661 \\
\quad 3D U-Net (det.)$^\dagger$ & 0.0762 & 0.1088 & 0.0981 & 0.0881 & 0.1025 \\
\quad 3D CVAE & 0.3450 & 0.4169 & 0.3454 & 0.3311 & 0.3484 \\
\quad Dich. Gaussian & 0.1455 & 0.0661 & 0.0681 & 0.0619 & 0.0640 \\
\quad Coupled GLM & 0.0311 & 0.0321 & 0.0382 & 0.0334 & 0.0268 \\
\bottomrule
\end{tabular}

  \caption{All five conditioning rungs, all four families, all six arms.
  \emph{random} supplies a mismatched context and is the control, not a rung.
  For families A, C and D the target is the true clip, so monotone improvement
  from \emph{random} to \emph{global full local} is the evidence that an arm
  reads its conditioning; the 3D U-Net moves the wrong way on family D, getting
  worse as context is added. Family B is scored differently and is exempt: it
  measures agreement with the descriptor the model was \emph{handed}, so an
  obedient arm matches whichever descriptor it receives and is flat across the
  ladder, and a high score under \emph{random} is the expected behavior rather
  than a failure to condition. The informative movement on B is the collapse at
  \emph{local only}, where our arm falls to $0.3081$: without the assay code it
  cannot act on the descriptor at all.}
  \label{tab:fullladders}
\end{table}

\section{Pooled marginal statistics}
\label{app:marginals}

The U-Net's win on marginal realism has a mechanical explanation. A
deterministic conditional-mean field, re-thresholded, sits closest to the
pooled marginal exactly when it is least committed to a particular clip,
which is why its score degrades from its own random-context null to full
context while ours improves.

\begin{table}[htbp]
  \centering
  \small
  \resizebox{\textwidth}{!}{\begin{tabular}{l r r r r r r r }
\toprule
statistic & 4C+soft (ship) & \multicolumn{2}{c}{MaskGIT-flat} & \multicolumn{2}{c}{DG (Macke'09)} & \multicolumn{2}{c}{GLM (Pillow'08)} \\
\cmidrule(lr){3-4}\cmidrule(lr){5-6}\cmidrule(lr){7-8}
 & median & median & $q$ & median & $q$ & median & $q$ \\
\midrule
firing rate & 0.2619 & 0.0523 & $1.3\times 10^{-143}$ & 0.0667 & $6.2\times 10^{-121}$ & 0.2365 & $7.0\times 10^{-3}$ \\
$4$-frame persistence & 0.7147 & 0.5278 & $9.1\times 10^{-20}$ & 0.4021 & $5.8\times 10^{-53}$ & 0.4512 & $7.2\times 10^{-49}$ \\
mean avalanche size & 0.4948 & 0.4602 & $2.8\times 10^{-1}$ & 0.3485 & $4.5\times 10^{-1}$ & 0.6832 & $5.0\times 10^{-34}$ \\
ISI distribution (KS) & 0.3465 & 0.5000 & $1.2\times 10^{-66}$ & 0.2055 & $2.0\times 10^{-91}$ & 0.2083 & $9.8\times 10^{-102}$ \\
avalanche size (KS) & 0.5000 & 0.5000 & $4.0\times 10^{-1}$ & 0.5000 & $7.0\times 10^{-6}$ & 0.6667 & $8.8\times 10^{-34}$ \\
aggregate & 0.4769 & 0.4404 & $3.7\times 10^{-11}$ & 0.3733 & $7.6\times 10^{-30}$ & 0.5138 & $1.5\times 10^{-1}$ \\
\bottomrule
\end{tabular}
}
  \caption{Per-clip relative error on the pooled summaries, median over clips,
  with the paired difference against ours corrected by BH within this family.
  Lower is better throughout. The matched peer wins the rate and persistence
  terms and we win the inter-spike-interval distribution; the two lookup arms
  win the terms a per-assay site map determines and lose the shape terms.
  These values are measured using the generation-regime sample sets, which
  contain four samples per clip. This protocol differs from that used in
  Table~\ref{tab:generation}, so the results are not interchangeable.}
  \label{tab:marginals}
\end{table}

Pooled statistics are relatively insensitive to conditioning: after aggregation
over clips, a model can match the correct marginal while assigning it to the
wrong clip. We therefore treat these statistics as complementary diagnostics and
base the conditioning claims on per-clip paired comparisons.

\section{Per-setting battery}
\label{app:battery}

\begin{figure}[htbp]
  \centering
  \includegraphics[width=\textwidth]{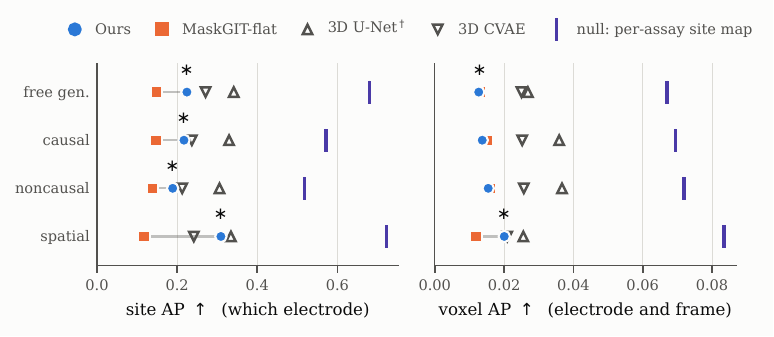}
  \caption{\textbf{Completion accuracy, paired per clip.} The left panels show
  site-level accuracy; the right panels additionally require the correct
  frame. $\ast$: the paired test favors
  us over the matched peer at $q<0.05$; the gray connector joins that pair.
  $\dagger$: no reusable representation (Appendix~\ref{app:baselines}). The
  violet rule is the static per-assay site map, a lookup that is constant
  within an assay and carries no clip-specific information. It still outranks
  every learned arm here. That is a property of average precision, which ranks a
  probability map and is therefore maximized by the conditional mean, not by a
  sample; generation is scored in Figure~\ref{fig:generation}. Panels have
  independent $x$ axes.}
  \label{fig:tasks}
\end{figure}

Figure~\ref{fig:tasks} and Table~\ref{tab:tasks} are the completion accuracy
behind Section~\ref{sec:tasks}, per setting and per arm.

\begin{table}[htbp]
  \centering
  \small
  \begin{tabular}{l r r r r}
\toprule
arm & free gen.\ (0) & causal (1) & noncausal (2) & spatial (3) \\
\midrule
\multicolumn{5}{l}{\textit{Site AP $\uparrow$: which electrode}} \\
Ours & 0.2244 & 0.2170 & 0.1889 & 0.3092 \\
MaskGIT-flat & 0.1481$^{\ast}$ & 0.1472$^{\ast}$ & 0.1387$^{\ast}$ & 0.1174$^{\ast}$ \\
3D U-Net$^\dagger$ & 0.3412 & 0.3297 & 0.3057 & 0.3342 \\
3D U-Net, seed 2$^\dagger$ & 0.4045 & 0.3414 & 0.3079 & 0.3475 \\
3D CVAE & 0.2709 & 0.2368 & 0.2127 & 0.2417$^{\ast}$ \\
\emph{null} assay site map, seen & 0.6796 & 0.5714 & 0.5184 & 0.7224 \\
\emph{null} assay site map, unseen & 0.0284 & 0.0230 & 0.0215 & 0.0350 \\
\midrule
\multicolumn{5}{l}{\textit{Voxel AP $\uparrow$: which electrode \emph{and} frame}} \\
Ours & 0.0127 & 0.0137 & 0.0154 & 0.0200 \\
MaskGIT-flat & 0.0130$^{\ast}$ & 0.0151$^{\circ}$ & 0.0162$^{\circ}$ & 0.0119$^{\ast}$ \\
3D U-Net$^\dagger$ & 0.0269 & 0.0359 & 0.0367 & 0.0255 \\
3D U-Net, seed 2$^\dagger$ & 0.0322 & 0.0339 & 0.0355 & 0.0253 \\
3D CVAE & 0.0250 & 0.0252 & 0.0257 & 0.0209 \\
\emph{null} assay site map, seen & 0.0670 & 0.0695 & 0.0719 & 0.0835 \\
\emph{null} assay site map, unseen & 0.0022 & 0.0021 & 0.0030 & 0.0034 \\
\bottomrule
\end{tabular}

  \caption{Average precision inside the hole, both levels, all four tasks.
  $\ast$: the paired test favors ours; $\circ$: not significant; unmarked:
  the other arm wins. Read within a column only: ROI fractions differ
  ($1.00/0.59/0.40/0.34$) and so do the clip counts
  ($279/278/278/243$), so the columns have different base rates. Nothing is
  bolded because the arms serve different roles and are not a single ranked
  benchmark.}
  \label{tab:tasks}
\end{table}

\paragraph{One readout for every arm.}
Every arm is scored through the same readout code path. An arm that emits a
score is thresholded; an arm that emits a probability is sampled. This matters
because an early version of the coupled GLM under-produced spikes by $4\times$
purely from a rate clamp in its own sampler, which is a property of the
baseline's implementation and not of the method, and could have been mistaken
for a property of the baseline. Own-count columns (accuracy given the true spike count) are reported for every task in the battery, because at this sparsity the count and
the placement are separable and an arm can win one while losing the other.

\paragraph{Free generation against reconstruction.}
Task~0 has an ROI of $1.000$: nothing is visible. It is therefore free
generation, and its output is bitwise identical for every arm regardless of
which clip was nominally being completed. It is included as the no-visible-volume reference, and it is not a fourth
completion task.

\begin{figure}[htbp]
  \centering
  \includegraphics[width=\textwidth]{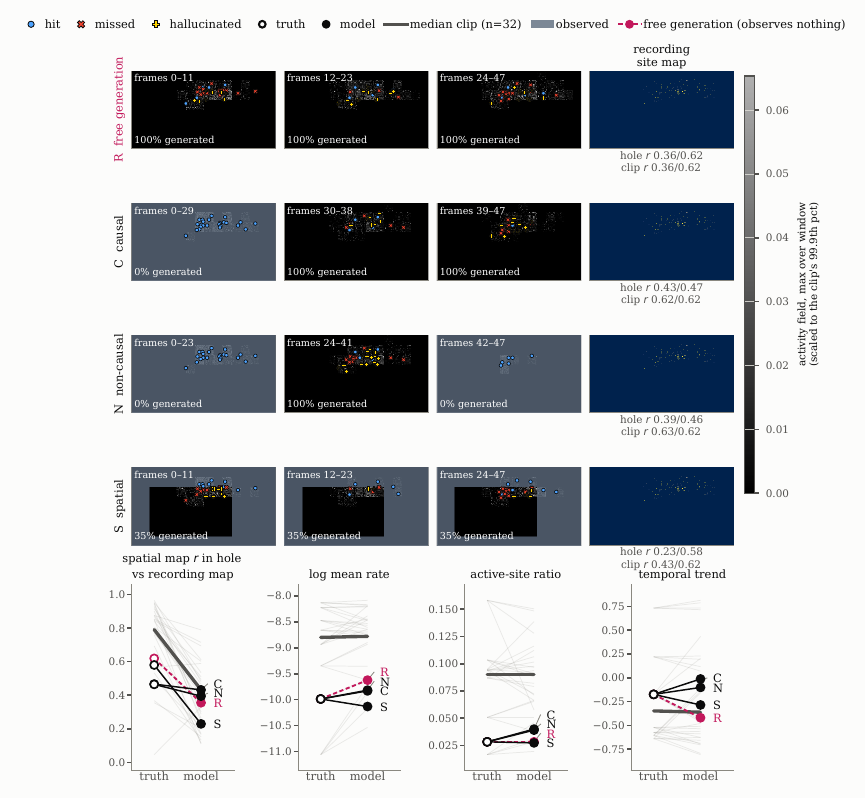}
  \caption{One ground-truth clip under all four settings. The dump indexes the
  same clip under every task, so these rows are one clip seen through the whole
  conditioning ladder and not four unrelated clips. The strip therefore shows
  a single hollow truth point connected to four corresponding model points.
  Washed regions are observed, not generated, and each panel states the
  fraction it produced; without that mark, spikes copied from the visible
  remainder could be mistaken for generated hits. The map column is measured inside the hole
  only. A whole-clip correlation would be symmetric (the copied remainder is identical in the model's clip and in the truth) but not comparable
  across these rows: for a temporal hole every site is observed in some frame,
  so projecting over time fills the canvas from copied frames, whereas under
  the spatial hole a fraction $0.352$ of sites is never observed at all. Inside
  the hole the attained fraction is $0.9258$ from a prefix and
  $0.8486$ with two-sided context, against $0.5741$ with nothing
  observed and $0.3954$ when the hole is a region of the array. Both
  inpainting settings clear the zero-context reference, so the visible frames
  are being used to place activity and not merely to fill the rest of the
  volume. Free
  generation is drawn dashed in crimson because it observes nothing and is
  therefore not a completion setting at all
  (Appendix~\ref{app:battery}); it is the zero-context reference with which the
  other three settings are compared, and the spatial setting is the only one that falls below
  it. The whole-clip value is printed under each site map for reference and
  runs $0.5741$--$1.0181$; the gap between the two columns is the copied share,
  not a result. The spatial setting is the weak one, and it
  is where the static site map of Appendix~\ref{app:battery} beats us.}
  \label{fig:qual_gen}
\end{figure}

Figure~\ref{fig:qual_gen} shows one clip completed under all four settings,
with the observed remainder marked so that copied spikes are not read as
generated ones.

\section{Qualitative panels}
\label{app:qualitative}

Both figures here are selected examples, so the selection rule is stated and
the distribution it selected from is drawn alongside. Panels are maximum
projections over a $16$-frame window, not single frames: at this
occupancy a frame carries a handful of spikes over $26{,}880$ sites, so an
instant is visually empty and shows nothing about motif structure. Rows are
chosen by the fraction of \emph{achievable} spatial-map correlation the model
attains, $r_{\text{model}}/r_{\text{truth}}$, where the denominator is the
clip's own ground truth against its assay's training site map. Selection, the ranking it came from and every number quoted below are written to a committed artifact by the panel generator.

Spatial-map correlation carries the tolerance the rest of the pipeline
already uses. Placement is learned through the global code and a
spatial-violation term, never supervised per electrode, so both the training
loss and the generation metrics accept a prediction within one site of an allowed one, a dilation of radius $1$; the same radius is applied here, to the model map and to the reference alike so that
the ceiling moves with it. The reference is the clip's own ground truth
against the same training site map, and it is far below $1$ for a reason
that bounds any single-clip generator: the map is an average over many
clips, while one clip is a sparse draw from it, a few hundred spikes over
$26{,}880$ sites. Two real clips of the same assay agree only to a
median $r$ of $0.7282$ across the $31$ assays. Panels therefore report the
attained fraction, model over reference, not a raw correlation.

\section{Count discrimination and its arithmetic control}
\label{app:counts}

How many spikes a clip contains and where it puts them are separable at this
sparsity, so the battery reports both. Within an assay, our correlation with
the true ROI spike count is $0.9138$--$0.9638$ across the three completion
settings, against $0.0716$--$0.4892$ for MaskGIT-flat.

That result must be compared with an arithmetic control and not with zero.
Every model is given the clip's local code. It is computed on the whole clip,
including the hidden region, and that is deliberate: the completion tasks here
supply a description of the clip and ask the model to fill the gaps
consistently with it, so the descriptor is a prompt and not something the model
is expected to infer. A consequence is that its first feature is log mean
firing density, so $\exp(\mathrm{lct}_0)\times|\mathrm{ROI}|$ predicts the
count with no model at all and scores $0.4550$--$0.8328$. It provides a
model-free baseline that a useful count head should exceed, is computed
alongside the other model-free references, and is exceeded by our model in all
three completion settings.

Discrimination and calibration are separate capabilities and no arm here has
both, and the two count quantities this battery reports must be kept apart. The
own-count column sums the decoded voxel probabilities inside the hole and takes
that sum as the arm's expected spike count; ours runs about $6\times$ the true
count, and the only arm whose bias stays within $\pm25\%$ everywhere is a
per-assay lookup \citep{elkan2001costsensitive}. That figure belongs to the
decoded probability field, not to the activity prior's categorical count head,
which is a separate output. Measured on its own, the head is unbiased under the
random masks it trains on and over-predicts by $13.2\%$ on train clips and
$15.2\%$ on validation clips under the fixed protocol that validation and
generation use, whose distribution of hole sizes it never saw. One fitted
multiplicative scale on held-out train clips takes that bias to $2.4\%$ and
count MAE from $10.500$ to $8.913$; an affine fit and a hole-size covariate add
nothing over it. The shipped default leaves the scale at $1$, so no generation
number in this paper depends on the correction. Ranking the right sites and
emitting the right number of them are different problems. The count head sets
the second at token level: its logits give the $K$ active cells the Gumbel
top-$K$ readout draws. How many voxels that becomes is then set by the decoder
and the voxel readout, which is why the token-level bias above and the decoded
over-count are different numbers.

\section{Null construction}
\label{app:nulls}

A margin over uniform says very little at $V = 961$.
Table~\ref{tab:priornulls} therefore scores the motif prior against a ladder of
four progressively harder model-free lookups.

\begin{table}[htbp]
  \centering
  \small
  \begin{tabular}{l r r r r r}
\toprule
predictor & top-1 $\uparrow$ & top-5 $\uparrow$ & median rank $\downarrow$ & MRR $\uparrow$ & CE (nats) $\downarrow$ \\
\midrule
uniform & 0.0024 & 0.0044 & 481 & 0.0021 & 6.8680 \\
global marginal & 0.0078 & 0.0321 & 156 & 0.0305 & 6.2895 \\
per-assay marginal & 0.0149 & 0.0599 & 102 & 0.0497 & 6.3026 \\
per-assay $\times$ position & 0.0675 & 0.2512 & 53 & 0.1529 & 6.1713 \\
\midrule
\textbf{motif prior (ours)} & \textbf{0.0860} & \textbf{0.3668} & \textbf{9} & \textbf{0.2168} & \textbf{4.1296} \\
\bottomrule
\end{tabular}

  \caption{The motif prior against its four-rung null ladder over $V=961$.
  Median rank is quoted because the distribution has a long tail and the mean
  ($44.65$) misrepresents it. The ladder is shown in full so the margin is
  compared with its hardest rung, the per-assay code distribution at that
  grid position, and not with uniform. The MRR here is teacher-forced: the
  prior is scored against the true activity field. Table~\ref{tab:seeds}
  reports the deployment figure instead, under the activity field the prior
  itself emits, which is lower and is the number the shipped model is selected
  on.}
  \label{tab:priornulls}
\end{table}

The four rungs of Table~\ref{tab:priornulls} are cumulative lookups fitted on
the training split and applied unchanged at test time. \emph{Uniform} is $1/V$. \emph{Global marginal} is the
empirical code distribution pooled over all assays. \emph{Per-assay
marginal} conditions that distribution on the assay identity.
\emph{Per-assay $\times$ position} conditions it on the assay and the
grid cell jointly, and is the strongest: it is the best a pure lookup table can
do without seeing the clip.

\paragraph{The random rung as a control.}
The conditioning ladder in Appendix~\ref{app:ladders} has a \emph{random} rung
in which the model is given another clip's context. It is tempting to interpret the
correlation across all five rungs as a measure of conditioning strength, but
the random rung supplies an assay index that bypasses the ladder's ordering
entirely, so a pooled correlation over the five rungs measures the wrong thing.
The quantity we report is the within-assay gap between the full-context and
random-context rungs, which is paired and does not mix the two effects.

\section{Baseline implementations}
\label{app:baselines}

\paragraph{Stationary fits for both statistical baselines.}
A full per-electrode covariance over the $26{,}880$ padded canvas locations is
not estimable from the available windows at this firing rate, so neither a
per-channel covariance nor per-electrode coupling can be fitted. Both models are
therefore used at the stationary parameterization their own literature
prescribes; this is a property of the data, not a weakening of the baselines.

\paragraph{Free generation: mean against median.}
We win free generation on the paired test while posting a marginally lower
\emph{mean} voxel AP: we are ahead on $62\%$ of clips, and a few clips on which
the flat arm does much better move the mean without moving the median. The
paper therefore quotes the paired test.

\paragraph{Win-loss count across the paired comparisons.}
Across the two metric families and four settings, other methods outperform
ours in $23$ of the $32$ paired comparisons; $7$ go to us and $2$ are ties.
All $23$ wins are achieved by the directly supervised convolutional inpainters
(the U-Net, its seed replicate and the CVAE), and \emph{none} by the matched
generative peer, against which we take $6$ of $8$ with the other two tied.

\paragraph{The U-Net as an upper reference.}
It beats us on voxel AP in all four completion settings
($q \leq 3.53\times10^{-5}$), and it is deterministic: one context yields one
field, so every ``sample'' is that field re-thresholded. Its marginal-realism
score worsens as context is added (Section~\ref{sec:generation}), and it has no
entry in the alphabet comparison because it has no discrete codebook. The dichotomized Gaussian likewise emits a static per-site
field that cannot read the visible remainder.

\paragraph{Cost of withholding the site map.}
Replacing the per-assay map with a single global one takes GLM adherence
from $0.5012$ to $0.0188$ and DG's from $0.5472$ to $0.2076$, and DG's
within-assay placement gap falls fivefold, from $0.0058$ to $0.0011$.
Adherence at $0.0188$ means the GLM does not degrade so much as stop tracking
the requested context altogether; its own placement gap is negative both with
the map and without it, so that capability was never there to lose. This is the
measurement behind Section~\ref{sec:tasks}: what these arms score is the map.

\begin{table}[htbp]
  \centering
  \begin{tabular}{l r r r r}
\toprule
arm & fitted (shared) & shared maps & per assay & at 1000 assays \\
\midrule
Ours & 8,620,946 & 0 & \textbf{0} & 0 \\
MaskGIT-flat & 13,232,961 & 0 & \textbf{0} & 0 \\
3D U-Net$^\dagger$ & 8,372,673 & 860,160 & \textbf{0} & 0 \\
3D CVAE & 9,360,601 & 860,160 & \textbf{0} & 0 \\
\emph{ref} Dich.\ Gaussian & 26 & 26,880 & \textbf{26,881} & 26,881,000 \\
\emph{ref} Coupled GLM & 5,194 & 53,792 & \textbf{26,880} & 26,880,000 \\
\bottomrule
\end{tabular}

  \caption{Parameters shared across assays versus tabulated per assay.
  The per-assay column counts one value per padded canvas location, not per
  physical electrode and not per routed channel. The two lookup arms grow
  linearly in the number of assays; every learned arm, including the peer,
  is flat. At \NAssays\ assays that is $833$k tabulated values, and at a
  thousand it would be $26.9$M.}
  \label{tab:scalability}
\end{table}

Table~\ref{tab:scalability} is the parameter count behind that claim.

\paragraph{MaskGIT-flat.}
MaskGIT-flat uses a single codebook of $1024$ entries on the identical
$8\times8\times16$ grid and $(6,15,14)$ patch, together with the same prior
family and conditioning inputs. Token budget, alphabet size and training budget are
matched. Its tokenizer is the ordinary one this literature describes: strided-convolution patch embedding, four residual convolutional blocks, one
codebook, mirrored decoder, with no attention anywhere in it. It is built that way deliberately, since comparing against our own tokenizer with the
ladder switched off would not provide an independent architectural baseline. The
consequence is that the arm does \emph{not} isolate the residual ladder: it
varies the tokenizer core, the codebook structure, the blank route and the
auxiliary losses together. What it bounds is the pair, and the oracle-code
comparison in Section~\ref{sec:alphabet} removes only the prior from that
pair, not the encoder architecture.

\paragraph{3D U-Net and 3D CVAE.}
The U-Net is an inpainter: it takes the masked volume and the mask, and is
trained with the same loss on the same hole distribution. It has no
autoencoding path, so it cannot reconstruct, and its free-generation output is
the all-masked corner of the same task, driven by FiLM on the conditioning
codes and thresholded at a fitted log-linear rate. The CVAE is the same
backbone with a latent variable added and nothing else changed, a conditional VAE in the standard form \citep{sohn2015cvae}, with a recognition
network $q(z \mid x, c)$ used only in training, a \emph{conditional} prior
$p(z \mid c)$ and a generator $p(x \mid z, c)$, where $c$ is the visible
remainder, the hole mask and the two conditioning codes. It therefore samples
rather than regresses, and its free-generation output is a genuine draw from
that conditional prior, not a posterior mean.

Both required an explicit positional embedding: FiLM conditioning is spatially
uniform, so without one a convolutional stack cannot represent a per-assay
site map at all, and the U-Net's spatial-map correlation falls from $0.2757$ to
$0.0552$.

The CVAE's latent nevertheless collapses, and Table~\ref{tab:cvaecollapse}
records the measures used to prevent posterior collapse and the resulting KL.
These measures include a conditional prior, a down-weighted KL, KL warm-up,
per-channel free bits \citep{kingma2016iaf}, and charging KL only where the
hole is; two independent fits collapse despite these measures.

We interpret this collapse as consistent with the conditional structure of the
task rather than evidence, by itself, of an optimization failure. The conditional
prior cannot see inside the hole:
at generation time nothing about the held-out region is available to it, so
the skip connections already supply much of the available conditioning; the
ELBO therefore has little incentive to retain additional latent information. Forcing
usage with a larger $\beta$ floor or higher free bits would increase latent
activity without establishing that it carries useful held-out information. The consequence for the tables is that this
arm behaves like the deterministic one, which is why the two convolutional arms
move together throughout.

\begin{table}[htbp]
  \centering
  \small
  \begin{tabular}{@{}l r r r l@{}}
\toprule
fit & epochs & peak val KL/dim & final val KL/dim & collapsed \\
\midrule
shipped & 31 & 0.1300 & 0.0100 & yes \\
no positional embedding & 40 & 0.0270 & 0.0089 & yes \\
\bottomrule
\end{tabular}

\vspace{0.4em}

\begin{minipage}{\textwidth}\footnotesize
Both fits carry the standard anti-collapse measures: a \emph{conditional} prior $p(z \mid c)$ in place of $\mathcal{N}(0, I)$; the KL down-weighted to $\beta = 0.05$; KL warm-up over the first 30\% of training; free bits at 0.05 applied \emph{per channel}, which is where collapse happens; the KL charged only on the latent cells the hole touches, so the latent is not taxed for what the skip connections already carry; and the posterior log-variance initialized at $-4.0$. The latent is a 4-channel grid and not a global vector, so it can say \emph{where} the extra spikes go.
\end{minipage}

  \caption{The CVAE's latent, and the measures taken against collapse. Two
  independent fits, differing only in whether the backbone carries a positional
  embedding, both end below the free-bits floor. Settings and outcomes are read
  from the fit reports.}
  \label{tab:cvaecollapse}
\end{table}

\paragraph{Dichotomized Gaussian and coupled GLM.}
Both are fitted per assay at the stationary parameterization their own
literature prescribes. The data force this choice: at a rate
of \VoxelRate\ and roughly $200$ spikes per clip, a full per-electrode
covariance or a per-electrode GLM over the $26{,}880$ padded canvas locations
is not estimable from the available windows. Table~\ref{tab:withheldmap} is the
test that establishes what these two arms are actually using.

\begin{table}[htbp]
  \centering
  \small
  \begin{tabular}{l r r r r r r}
\toprule
\emph{ref} arm & \multicolumn{2}{c}{within-assay gap} & \multicolumn{2}{c}{adherence} & \multicolumn{2}{c}{map $r$} \\
\cmidrule(lr){2-3}\cmidrule(lr){4-5}\cmidrule(lr){6-7}
 & with & without & with & without & with & without \\
\midrule
Dich.\ Gaussian & +0.0058 & +0.0011 & +0.5472 & +0.2076 & 0.5807 & 0.0694 \\
Coupled GLM & -0.0049 & -0.0004 & +0.5012 & +0.0188 & 0.5883 & 0.0624 \\
\bottomrule
\end{tabular}

  \caption{The two lookup arms with their per-assay site map replaced by
  the global one, same models and same clips. The GLM's adherence does not
  merely degrade, it goes negative: without the map it stops tracking the
  requested context at all. In this evaluation the two arms therefore function
  primarily as assay-specific memorization references.}
  \label{tab:withheldmap}
\end{table}

\section{Seed variance}
\label{app:seeds}

The adaptation stage produces a modest improvement, so we test whether it
exceeds variation across seeds. Table~\ref{tab:seeds} reports four runs.

\begin{table}[htbp]
  \centering
  \small
  \begin{tabular}{l r r r r}
\toprule
run & best epoch & epochs & val MRR & deployment MRR \\
\midrule
\textbf{shipped} & 24 & 64 & 0.21359 & 0.20351 \\
seed 101 & 56 & 96 & 0.21254 & 0.20581 \\
seed 202 & 35 & 75 & 0.21225 & 0.20447 \\
seed 303 & 80 & 120 & 0.21377 & 0.20649 \\
\midrule
mean $\pm$ sd & & & 0.21304 $\pm$ 0.00076 & 0.20507 $\pm$ 0.00134 \\
\midrule
\emph{ref} no adaptation, soft field & & & & 0.18981 \\
\emph{ref} no adaptation, hard map & & & & 0.14266 \\
\emph{ref} oracle activity map & & & & 0.21488 \\
\bottomrule
\end{tabular}

  \caption{The adaptation stage across four independent seeds, each
  re-initialized from the unadapted control and run through the real dispatch
  path. The effect over the soft-field reference is $+0.01526$ against an
  across-seed sd of $0.00134$, i.e.\ $11.4\times$, and the \emph{worst} seed
  still beats that reference by $+0.01370$. The selected epoch varies from $24$ to
  $80$ while the value moves $0.0015$, so early stopping is picking a plateau
  and not a lucky checkpoint.}
  \label{tab:seeds}
\end{table}

Table~\ref{tab:seeds} also shows two important features of checkpoint variability. The shipped checkpoint is the
\emph{worst} of the four on deployment MRR. It was selected on validation MRR,
and validation turns out to be a poor predictor of deployment across seeds: seed $101$ has the worst validation score and the second-best deployment score.
Swapping in seed $303$ after seeing the test numbers would be selecting on
test, so the val-selected checkpoint ships and the penalty is reported. And the
same seed test settles the direction of adaptation.

\paragraph{Direction of the adaptation stage.}
The reverse direction, which freezes the motif prior and tunes the activity
prior on the downstream objective, faces three limitations. The sampled map is discrete,
so the gradient is a straight-through surrogate through a top-$K$ whose true
Jacobian is zero almost everywhere; the trainable surface cannot change the
\emph{ranking} that decides which cells light; and ``make the map closer to
truth'' is already the activity prior's own training objective, pursued with
full gradient access. Measured over $56$ epochs, every candidate sat inside one seed
standard deviation, with motif MRR declining at $t = -10.45$. That entire range
inside $0.92$ seed sd distinguishes it from the shipped direction, whose effect
is $11.4$ times the same quantity.

\section{Limitations: supporting detail}
\label{app:limits}

\paragraph{The nature of the input.}
The model consumes a binary volume over the array footprint, but that volume is
built from sorted units: each curated unit is written to its peak electrode as
a point event (Appendix~\ref{app:preproc}). So the input is a spatial raster of
sorted unit locations, not raw threshold crossings and not unsorted multi-unit
activity. What the phrase \emph{array-wide binary volume} claims is about the
model and not the recording: there is no per-unit parameter, no unit
correspondence across assays, and no unit-indexed output head. Sorting acts as
an upstream filter that decides which events enter the canvas.

\paragraph{Scope of the forward model.}
Every clip here is spontaneous activity. Nothing in the corpus carries
electrical stimulation, and the conditioning interface has no input for one, so
the model gives the distribution of activity a preparation produces when left
alone. That supports simulation and supplies the unperturbed baseline against
which a stimulus-evoked change would be measured. It does not by itself support
offline controller design, which needs transition dynamics conditioned on
applied stimulation; reaching that would mean extending the conditioning
interface to carry stimulus events and training on recordings that contain
them.

\paragraph{Resolution and its uses.}
The prior recovers $80$--$98\%$ of its alphabet's site-level oracle-code
reference and $4$--$6\%$ of the voxel-level one, and the within-token temporal
profile is close to uniform (Appendix~\ref{app:blur}). The model should
therefore be read as generating which electrodes are active and the envelope of
a burst, not the millisecond placement of spikes within an active patch. Uses
that turn on exact spike timing are outside what the present resolution
supports.

\paragraph{Recordings, preparations and memorization.}
The organoid assays come from a cohort of at most six organoids and the archive
does not say which assay came from which; the slice assays come from two slice
preparations, one per neurosurgical patient. Generalization to an unseen
preparation is not only untested but impossible under the current conditioning
interface: the per-assay code is a
seeded random vector, so a new preparation supplies the model with no
information (Section~\ref{sec:tasks}). Assay-specific information
certainly does live in the shared weights. The claim is not that the model
has memorized nothing, only that what it memorizes does not grow a table as
assays are added.

\paragraph{Smaller patches.}
They reconstruct better but leave $97\%$ of tokens blank, and a prior trained on
a grid that empty collapses onto predicting blank. The patch size is therefore a
tokenizer--prior trade-off resolved in favor of the prior, not an optimum for
either alone.

\paragraph{Exclusion of the blank token from the reuse statistics.}
It is $91.7\%$ of all tokens and every assay emits it, so including it would
push every pairwise overlap toward $1.0$ while measuring nothing. Excluding it
therefore makes the overlaps reported in Section~\ref{sec:reuse} conservative
and not artificially favorable.

\paragraph{Token granularity.}
Our alphabet's site-level oracle-code reference on free generation is
\OracleOursSite\ and the prior reaches $0.2244$, $98\%$ of it. The voxel-level
reference is \OracleOurs\ and the prior reaches $0.0127$, $5\%$ of it. The gap
between those two fractions quantifies the within-token timing problem:
the model places activity on very nearly the right electrodes and then
distributes it wrongly in time inside the $36$\,ms token.

\paragraph{Evidence on the tokenizer core.}
A convolutional VQ tokenizer is measured here: MaskGIT-flat's is fully
convolutional, with no attention anywhere in it, on the same patch and grid at
a matched token budget, and given the true codes it represents held-out
activity a factor of \CeilingFactor\ worse than ours
(Section~\ref{sec:alphabet}). That is the closest evidence available, and it is
not a controlled comparison: the same arm also replaces the three-level ladder
with one flat codebook and drops the blank route, so core and alphabet vary
together.

A second bound comes from the U-Net, which reaches spatial-map correlation
$0.2757$ with a positional embedding and $0.0552$ without, against our $0.3050$.
The positional-embedding ablation is consistent with the expected mechanism:
convolution is translation-equivariant whereas the electrode array has absolute
site identity, so an explicit spatial basis is needed to represent a fixed site
map. It bounds how much of our spatial result is attributable to the
transformer and how much to the patch grid and the positional information, but
the U-Net differs from our tokenizer in objective and supervision as well as in
structure.

\section{Within-token blur}
\label{app:blur}

A $(6,15,14)$ patch is $1260$ voxels and the decoder places probability across
it too evenly. Over content tokens, the model's within-token
temporal profile entropy is $1.7226$ nats against a real-data value of $0.3282$
and a completely flat ceiling of $1.7918$; spatially it is $3.8674$ against
$0.2478$ with a ceiling of $5.3471$. Two fixes were tried and both are reported
as failures.

A \emph{peak term} rewarding concentration within the token converges only as a
late fine-tune; from scratch it prevents the tokenizer converging at all. As a
fine-tune it sharpens all four profile axes, but at a substantial cost in AP.
A \emph{token-entropy constraint} on the within-token profile is dose-ordered
and also too expensive: at the higher of two doses it moves the temporal profile
entropy from $1.7801$ in the matched unconstrained arm to $1.7622$ and the spatial from $4.4521$ to $4.2710$ ($1.2\%$ of the temporal gap to real data and $4.3\%$ of the spatial) for a $28\%$ fall in validation AP as the ramp
completes.

All three numbers are read from the \emph{final} checkpoint and not the
validation-selected one. The term is ramped in late, so validation selected a
checkpoint three epochs after activation, at which point every diagnostic still
showed the untreated model; a ramped-in term measured anywhere but the last
epoch will appear to do nothing.

Neither fix is in the shipped model; the tolerant spike loss remains the only
blur control. The blur is why Section~\ref{sec:limitations} names within-token
timing, not spatial placement, as the dominant residual error.

\section{Stage 3: the local-context mapper}
\label{app:stage3}

\paragraph{The nine local-context scalars, and the level-1 basis.}
They are log mean firing density; the second moments of the spike mass in $x$,
$y$ and $t$ together with their three cross-terms; the active-site ratio; and a temporal trend score. All are computed directly from the volume, with nothing learned. The clustering target is a $128$-texton basis built on the first ladder
level rather than on the flat alphabet because level~1 carries motif identity,
while levels~2 and~3 carry fine-grained details that the nine clip-level statistics cannot
predict.

\begin{table}[htbp]
  \centering
  \small
  \begin{tabular}{l r r r r}
\toprule
conditioning & $\Delta$NLL flat $\uparrow$ & $\Delta$NLL texton $\uparrow$ & $R^2$ $t$-centroid $\uparrow$ & $R^2$ $t$-marginal $\uparrow$ \\
\midrule
\texttt{lct} only & 0.2072 & 0.0943 & 0.1156 & 0.9027 \\
\texttt{gct} only & 0.3036 & 0.0861 & 0.0224 & 0.6063 \\
both & 0.2999 & 0.1017 & 0.1204 & 0.8885 \\
\bottomrule
\end{tabular}

  \caption{The three conditioning arms at epoch $300$, sharing one frozen
  encoder pass so the unique contribution of each is measured exactly. The
  local code carries temporal structure the global one cannot; the global code
  carries code identity the local one cannot; neither alone suffices. The
  $t$-centroid column is a negative result and is reported as one.}
  \label{tab:stage3}
\end{table}

Table~\ref{tab:stage3} separates what each code contributes. Convergence is not
assumed: the mean of the last $20$ epochs matches the final
epoch to four decimal places on every reported quantity. The base NLL against
which $\Delta$ is measured is $6.2482$ nats on the flat alphabet and $4.8164$
on the texton basis.

\paragraph{Multinomial likelihood against histogram $R^2$.}
Scoring a per-clip code histogram with $R^2$ has a hard ceiling that has
nothing to do with the model. Writing the target as a true rate plus multinomial
noise from drawing $n \approx 86$ active tokens gives
$R^2_{\max} \approx n\,\mathrm{CV}^2 / (V + n\,\mathrm{CV}^2)$, which is
$0.082$ at $V = 961$. More training approaches that ceiling and cannot move it.
A multinomial likelihood has no such denominator, because each clip contributes $n$ real draws, so a $961$-way alphabet is exactly as estimable as a $32$-way
one, and the score is also the objective the prior itself optimizes.

\begin{table}[htbp]
  \centering
  \small
  \begin{tabular}{l r r r r}
\toprule
basis, \#textons & base NLL & $\Delta$\texttt{lct} & $\Delta$\texttt{gct} & $\Delta$both \\
\midrule
$z_1$, 32 & 3.4221 & 0.0736 & 0.0716 & 0.0826 \\
$z_1$, 128 & 4.7935 & 0.0836 & 0.0808 & 0.0941 \\
$z_1{+}z_2$, 32 & 3.4162 & 0.0666 & 0.0628 & 0.0737 \\
$z_1{+}z_2$, 128 & 4.6878 & 0.0712 & 0.0683 & 0.0800 \\
flat, 32 & 3.4108 & 0.0607 & 0.0580 & 0.0672 \\
flat, 128 & 4.6948 & 0.0675 & 0.0643 & 0.0756 \\
\bottomrule
\end{tabular}

  \caption{Which basis the textons are clustered on, and how many. The ordering
  $z_1 > z_1{+}z_2 > \mathrm{flat}$ is monotone for all three arms, and $128$
  textons beat $32$ throughout. Descriptor space and decoder space serve
  different roles: level~1 carries reusable motif identity, while levels~2
  and~3 carry fine-grained details that the nine clip-level statistics cannot
  predict.}
  \label{tab:textonbasis}
\end{table}

Table~\ref{tab:textonbasis} reports the sweep over basis and texton count.

\section{Stage 1: the global-context pretrain}
\label{app:stage1}

The per-assay code is pretrained against that assay's spatial support map (which electrodes are ever active) so that the code acquires the
association before any prior consumes it. Per-assay support maps and adjacency diagnostics for all \NAssays\ assays are in the code release.

\paragraph{The conditioning interface.}
All assay-level conditioning enters as one vector per assay. The motif
prior reads it through the frozen Stage-1 mapper; the activity prior projects it
directly with a linear layer. Substituting a measured descriptor of a preparation (unit map, ISI distribution, days \emph{in vitro}, cell type, stimulation protocol) for the seeded random code is therefore a change to
those two input projections and nothing downstream.
We state this as a design property, not as a result: no experiment in this
paper substitutes anything, and the random code licenses no claim about a
preparation the model has not trained on.

\section{Reproducibility}
\label{app:repro}

The  code  release contains the model, all
training stages, all six evaluation arms including ours, the evaluation
harness, and the generators that produce every table and figure in this paper.
Its checkpoint manifest lists every checkpoint, marks which is shipped for each
stage, and records how it was selected. The README gives the ordered sequence
of commands the generators are driven by: the four extraction steps that write
the provenance, preprocessing, hyperparameter and evaluation-budget artifacts,
then the table and figure generators, then the prose check.  

Figures are byte-stable across runs. Every table is a generated file. Editing
one by hand would be overwritten and, worse, would silently disagree with the
run that produced it, so the generator or source artifact should be modified
instead. Numeric values in prose can become stale without being detected, and
the last step addresses this risk: it extracts every numeric literal from the
manuscript and fails if
one no longer appears in any committed artifact. The evaluation protocol is pinned as described in
Appendix~\ref{app:metrics}.

\end{document}